\documentclass[10pt,twocolumn,letterpaper]{article}
\usepackage[accsupp]{axessibility}
\usepackage{wacv}              

\usepackage{graphicx}
\usepackage{amsmath}
\usepackage{amssymb}
\usepackage{booktabs}
\usepackage{amsthm} 

\usepackage{algpseudocode}
\usepackage{algorithm}
\algnewcommand{\LineComment}[1]{\State \(\triangleright\) #1}

\newtheorem{prop}{Proposition}

\usepackage{lineno}
\usepackage{multicol}
\usepackage{lipsum}

\usepackage[pagebackref,breaklinks,colorlinks]{hyperref}

\usepackage[capitalize]{cleveref}
\crefname{section}{Sec.}{Secs.}
\Crefname{section}{Section}{Sections}
\Crefname{table}{Table}{Tables}
\crefname{table}{Tab.}{Tabs.}

\def\wacvPaperID{2082} 
\def\confName{WACV}
\def\confYear{2025}

\begin{document}

\title{From Visual Explanations to Counterfactual Explanations with Latent Diffusion}

\author{Tung Luu,  Nam Le,  Duc Le,  Bac Le\\
Faculty of Information Technology, University of Science, VNU-HCM\\
Vietnam National University, Ho Chi Minh City\\
{\tt\small \{20120616, 22C11067\}@student.hcmus.edu.vn, lnduc.xai@gmail.com, lhbac@fit.hcmus.edu.vn}
}
\maketitle

\begin{abstract}
Visual counterfactual explanations are ideal hypothetical images that change the decision-making of the classifier with high confidence toward the desired class while remaining visually plausible and close to the initial image. In this paper, we propose a new approach to tackle two key challenges in recent prominent works: i) determining which specific counterfactual features are crucial for distinguishing the ``concept'' of the target class from the original class, and ii) supplying valuable explanations for the non-robust classifier without relying on the support of an adversarially robust model. Our method identifies the essential region for modification through algorithms that provide visual explanations, and then our framework generates realistic counterfactual explanations by combining adversarial attacks based on pruning the adversarial gradient of the target classifier and the latent diffusion model. The proposed method outperforms previous state-of-the-art results on various evaluation criteria on ImageNet and CelebA-HQ datasets. In general, our method can be applied to arbitrary classifiers, highlight the strong association between visual and counterfactual explanations, make semantically meaningful changes from the target classifier, and provide observers with subtle counterfactual images. 
\end{abstract}
\section{Introduction}
\label{sec:introduction}

\begin{figure}
    \centering
    \includegraphics[width=1.0\linewidth]{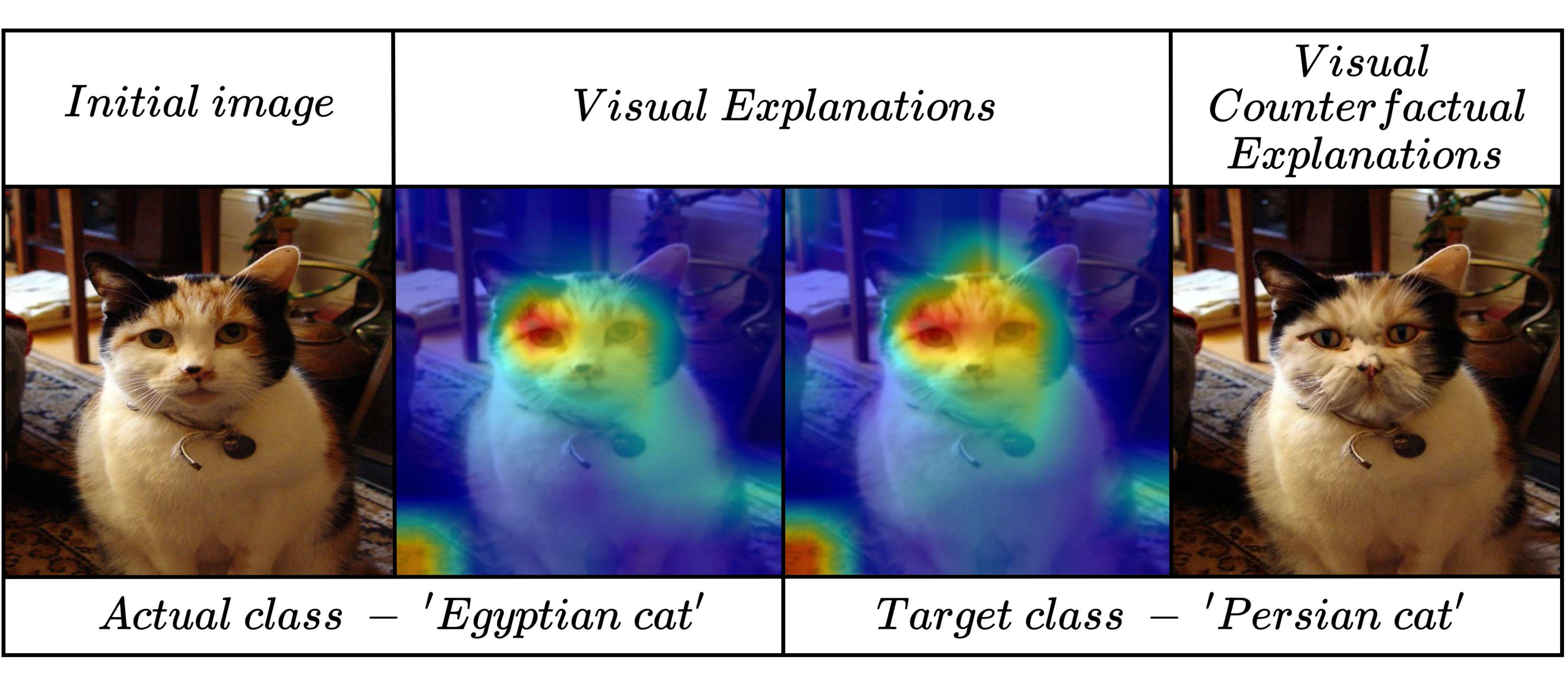}
    \caption{Illustration of local explanation techniques.}
    \label{fig:ce-vce}
    \vspace{-4mm}
\end{figure}

Local explanation methods enhance the ``transparency'' of the decisions of classification models for observers by identifying the significant elements of each input sample that influence the considered label. Recently, the class activation mapping method (CAM) \cite{zhou2016learning} and several variants \cite{selvaraju2020grad, wang2020score, wang2020ss, naidu2020cam} have highlighted key regions that a classification model relies on to identify a specific category. Nevertheless, a significant drawback of these visual explanation techniques is that they may highlight overlapping areas when explaining couples of labels with high similarity, making it difficult to identify the features distinguishing these categories. Counterfactual Explanations offer promising approaches to address this issue by generating hypothetical images that indicate the unique features of the target label compared with the original label. These explanations evoke causal reasoning in humans. The example illustrated in Figure~\ref{fig:ce-vce} demonstrates that the visual explanation method highlights features around the face for both labels, whereas our counterfactual approach indicates that: ``If this cat had longer and fluffier fur, it would be identified as a Persian cat instead of an Egyptian cat.", which did not previously exist. 

Recent cutting-edge studies have frequently integrated generative models into their proposed methodology to offer counterfactual explanations that align with human observational knowledge. However, providing high-quality hypothetical images still presents challenges, as some details that differ from the input sample lead observers to doubt whether these changes are necessary to achieve the desired label. Furthermore, these details mislead observers into believing that the classification model relies on these factors for its decision-making. Our study and ACE \cite{jeanneret2022diffusion} pursue the same objective of retaining essential details related to the desired label. When comparing the differences of pre-counterfactual explanations generated by diffusion models \cite{dhariwal2021diffusion, sohl2015deep, ho2020denoising, song2020denoising} with the input sample, ACE method can mark important regions that are misaligned if this region differs significantly more from the truly explanatory region for the desired label. For example, the context unrelated to the target label undergoes content or style-transfer changes due to the absence of masking techniques during the denoising process, while small adversarial gradients from the target model (e.g., the total magnitude of near-zero gradients) added to the image are insufficient to produce meaningful and significantly distinct changes. This challenge is discussed in several works on Adversarial Machine Learning \cite{moosavi2016deepfool, papernot2016limitations, papernot2016transferability, madry2017towards}, noting that a small perturbation across the entire image may alter the original label, even if the content of the image is visually almost unchanged. DVCE \cite{augustin2022diffusion} dealt with the above obstacle by leveraging adversarially robust models to make semantically meaningful changes. Observers aim to assess the knowledge capture capabilities of the target classifier; however, the supporting models may obscure their contributions.

In this study, we propose a novel framework, namely ECED, by investigating the affiliation between Visual \textbf{E}xplanations and \textbf{C}ounterfactual \textbf{E}xplanations, simultaneously utilizing Latent \textbf{D}iffusion Models to address the aforementioned challenges. Initially, our method determine minimal areas, defined as the foreground, in the original image that require modification using visual explanation techniques based on the provided label. To construct counterfactual explanations under the desired outcome, we leverage the blended latent diffusion algorithm introduced in \cite{avrahami2023blended}. Nevertheless, the drawback of this blending method is the requirement to optimize the decoder after each generating step to reconstruct the remaining part of the image, which is defined as the background. To address this limitation, we simultaneously fine-tune the decoder's weights and optimize the subspace of image embeddings to minimize content loss. This optimization focuses specifically on preserving the background of the reconstructed image compared to the initial one. In the final step, our method generates the counterfactual explanation by combining a blended diffusion process and adversarial attacks based on a pruning strategy. We also mitigate the loss of background content in the reconstructed image compared to the original throughout by reutilizing the components optimized in the preceding step.

In summary, the main contributions of this work are described as follows:
\begin{itemize}
    \item Through exploring the strong affiliation between visual and counterfactual explanations, our proposed ECED pre-defines the minimal regions required for changes, thereby limiting the appearance of unnecessary details in visual counterfactual explanations.
    \item We propose an enhancement strategy for the blended latent diffusion algorithm, a widely used masking technique applied to diffusion models, to adapt it for foreground editing tasks across multiple iterations with a low time cost while minimizing the loss of background content compared with the original.
    \item Our work makes semantic changes without using a robust classifier by boosting the target classifier-based adversarial attacks on latent space representing the foreground.
    \item ECED outperforms the previous state-of-the-art methods on most evaluation protocols across ImageNet and CelebA-HQ. Our method also provides counterfactual explanations that approximate human observational knowledge.
\end{itemize}

\section{Related Work}
\label{sec:related_work}

\subsection{Visual Counterfactual Explanations}
Generative networks such as GANs \cite{goodfellow2020generative} have been integrated into various counterfactual explanation studies (e.g. ExplainGAN \cite{samangouei2018explaingan}, and C3LT \cite{khorram2022cycle}). Diffusion models \cite{sohl2015deep, ho2020denoising, song2020denoising} have outperformed GANs and have been applied in numerous works, among which we would like to draw attention to two recent prominent studies: ACE \cite{jeanneret2023adversarial} and DVCE \cite{augustin2022diffusion}. ACE initially performs a denoising process on the entire pixel space based on DDPM \cite{ho2020denoising} along with adversarial attacks for several iterations to obtain a pre-explanation image. Then, by leveraging the RePaint algorithm \cite{lugmayr2022repaint}, this approach sparsely modifies the most distinct regions compared with the original image, concurrently reconstructing the content in the remaining areas. Our approach differs from ACE by pre-defining regions that are considered sufficiently small and significant to perturb, and then applying image masking to the diffusion model during denoising. DVCE \cite{augustin2022diffusion} addressed a key challenge of generating semantically meaningful changes by leveraging a pre-trained adversarially robust classifier. When the angle between the unit gradient of an adversarially robust classifier and that of the target classifier exceeds a permissible threshold, DVCE aligns the unit gradient of the target model with the robust model using cone projection. If the direction of these gradients differs significantly, the counterfactual explanation is generated primarily on the unit gradient of an adversarially robust model. In this work, we avoid using robust models and focus on tackling the difficulty emphasized by DVCE through gradient pruning. Our approach eliminates background-related adversarial gradient components, thereby increasing the number of adversarial attack iterations. As a result, a substantial amount of gradients of the target classifier is introduced into the latent features that represent the foreground.

\subsection{Blended Latent Diffusion}
\label{sec:blended}
In this work, we adapt Blended Latent Diffusion \cite{avrahami2023blended}, which is the image masking for text-to-image Stable Diffusion \cite{wang2018high}, in order to generate realistic counterfactual images while still preserving the background. Initially, the image $\mathbf{\mathcal{I}}^\textrm{F}$ is encoded into the latent space $\boldsymbol{z}^\textrm{F}$ through the encoder $\mathcal{E}(\cdot)$ of variational autoencoders (VAE) \cite{kingma2013auto}. Then this latent is represented as the input during the diffusion processes rather than an image in a high-dimensional space. The background and foreground contents of an image are represented by the subspaces of the latent features $\boldsymbol{z}^\textrm{bg}$ and $\boldsymbol{z}^\textrm{fg}$, respectively. At each timestep $t$, we obtain the state $\boldsymbol{z}^\text{bg}_t$ by directly noising the initial latent $\boldsymbol{z}^\textrm{F}$ with the specific Gaussian noise $\boldsymbol{\epsilon}_t \sim \mathcal{N}(0,\textbf{I})$, while the state $\boldsymbol{z}^\text{fg}_t$ is determined by denoising the subsequent state $\boldsymbol{z}_t$ with predicted noise $\boldsymbol{\epsilon}_\theta(\boldsymbol{z}_t,\mathbf{C}^\textrm{CF})$. Following the forward and backward processes introduced in DDPM \cite{ho2020denoising} and applying the reparameterization trick, latent representations are sampled at an arbitrary timestep $t$ as:
\begin{align}
    \boldsymbol{z}^\textrm{bg}_t &= \sqrt{\bar\alpha_t}\boldsymbol{z}^\textrm{F} + \sqrt{1 - \bar\alpha_t}\boldsymbol{\epsilon}_t, \label{eq:1}\\
    \boldsymbol{z}^\textrm{fg}_{t} & \approx \sqrt{\alpha_{t-1}}\hat{\boldsymbol{z}}_0 + \beta_{t-1}\boldsymbol{\epsilon}_\theta(\boldsymbol{z}_t, C^\textrm{CF}) + \sigma_t\boldsymbol{\epsilon}_t.\label{eq:2}
\end{align}
Please consult the Supplementary Material (Supp for short) for the implementation details and hyperparameter configurations of Equation~\ref{eq:1},~\ref{eq:2}. With downsampled binary mask $\mathbf{\mathcal{M}}'$, latent representation $\boldsymbol{z}_0$ is obtained through progressive denoising from $t = T_1$ to $t = 0$:
\begin{align}
    \boldsymbol{z}_t = \boldsymbol{z}^\textrm{bg}_t\odot(1-\mathbf{\mathcal{M}}') + \boldsymbol{z}^\textrm{fg}_t\odot\mathbf{\mathcal{M}}',
\end{align}
where $\{ \beta_t \}^{T_1}_{t=0}$ defines a linear noise scheduling, $\{ \sigma_t \}^{T_1}_{t=0}$ is the sequence of different variance hyperparameters, $\alpha_t = 1 -\beta_t$, $\bar\alpha_t = \prod_{s=0}^t \alpha_s$. Once the final latent $\boldsymbol{z}_0$ is reached, this noise-free latent is transformed to pixel space through the VAE Decoder $\mathcal{D}(\cdot)$. $\mathbf{C}^\textrm{CF}$ represents the text embeddings of the condition from a CLIP text encoder \cite{radford2021learning}. To guarantee the reconstruction of the original background and maintain the generated foreground of new image $\mathbf{\mathcal{I}}^\textrm{edit}=\mathcal{D}(\boldsymbol{z}_0)$, Avrahami~\textit{et al.}~\cite{avrahami2023blended} fine-tuned the decoder weights $\theta_\mathcal{D}$ to minimize pixel deviations:
\begin{align}
    \theta^* = &\operatorname*{arg\,min}\limits_{\theta} (\lVert \mathcal{D}_\theta(\boldsymbol{z}_0)\odot\mathbf{\mathcal{M}} - \mathbf{\mathcal{I}}^\textrm{edit}\odot\mathbf{\mathcal{M}} \rVert\nonumber\\
    & + \lambda\lVert \mathcal{D}_\theta(\boldsymbol{z}_0)\odot(1-\mathbf{\mathcal{M}}) - \mathbf{\mathcal{I}}^\textrm{F}\odot(1-\mathbf{\mathcal{M}}) \rVert).
\end{align}

\subsection{Class Activation Mapping}
CAM-based approaches \cite{chattopadhay2018grad, selvaraju2016grad, jiang2021layercam} produce a heatmap that emphasizes the pixels in the given image. The heatmap assigns greater temperature values to pixels that play a critical role in activating the model for recognizing a particular class. The fundamental concept for obtaining a CAM \cite{zhou2016learning} involves a linear combination of activation maps from the specific convolution layer. Grad-CAM \cite{selvaraju2020grad} utilizes the weights that correspond to the given class in the fully connected layer of the pretrained classifier as parameters of attention maps. Because the coefficients are fixed, this limitation may introduce bias by disproportionately assigning higher weights to certain activation maps. Score-CAM \cite{wang2020score} addressed this issue by dynamically calculating the parameters based on the influence of saliency maps for each input sample. This algorithm has been enthusiastically embraced by the community due to its exceptional performance in comparison with previous state-of-the-art works.
\begin{figure*}[ht!]
  \centering \includegraphics[width=0.7\textwidth]{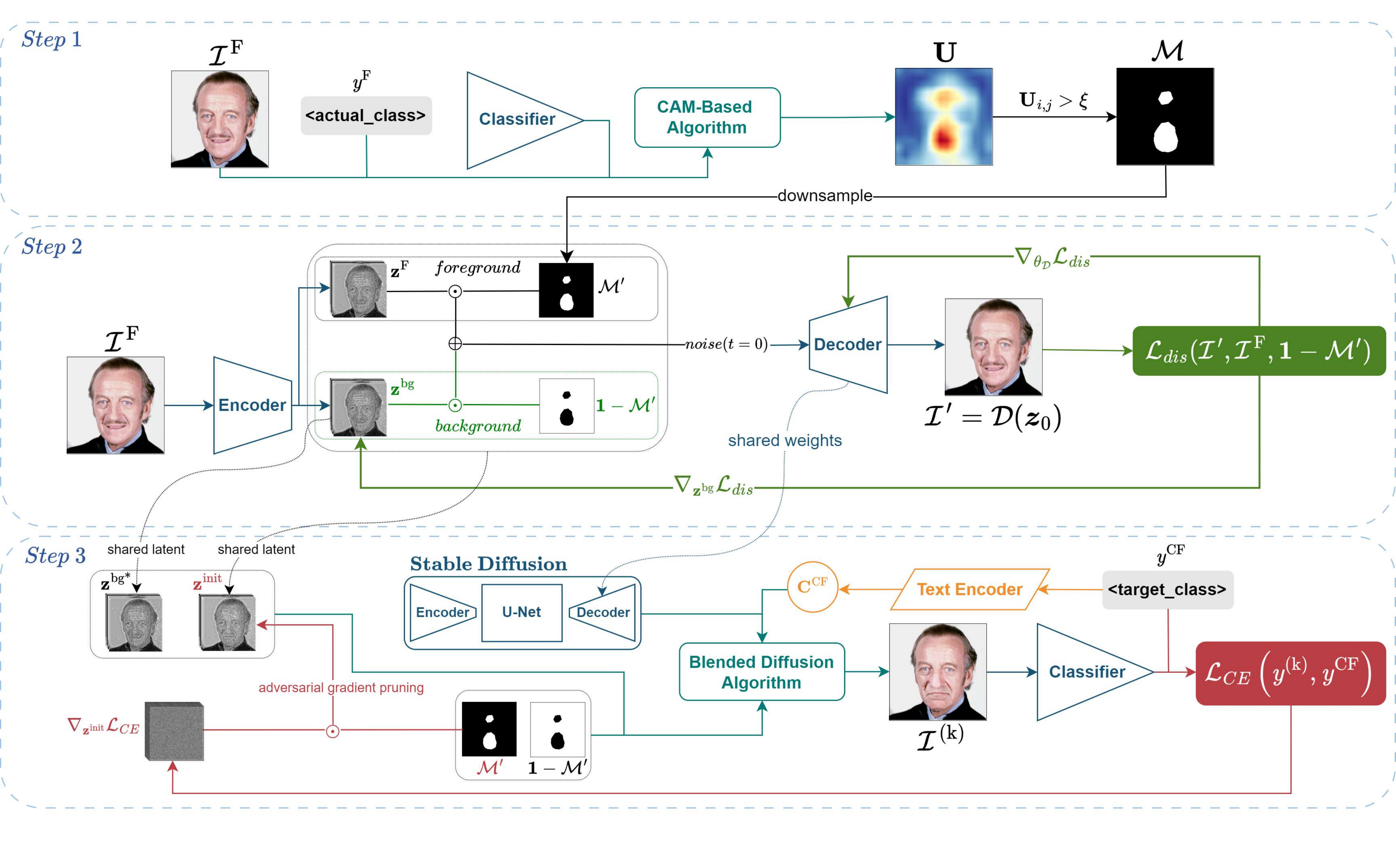}
  \vspace{-6mm}
  \caption{Our ECED generates counterfactual explanations in three sequential steps. Our method separates the foreground and background of the factual image by identifying a binary mask in first stage. Our method maintains the background content by using the fine-tuned VAE Decoder's weights and the optimized latent background while coming up with counterfactual explanations. Latent $\boldsymbol{z}^\textrm{init}$ is initialized by blending the original latent $\boldsymbol{z}^\textrm{F}$ and the optimized latent $\boldsymbol{z}^\textrm{bg*}$. Finally, we obtain counterfactual image $\mathbf{\mathcal{I}}^\textrm{CF}$ when $\mathbf{\mathcal{I}}^\textrm{CF} = \mathbf{\mathcal{I}}^{(T_2)}$.}
  \vspace*{-1.85mm}
  \label{demo_framework}
\end{figure*}
\section{Methodology}
\label{sec:methodology}
We propose the ECED framework, a novel approach to generate minimal and semantically meaningful changes, restricting redundant modifications while ensuring high fidelity to the desired label. Additionally, we eliminate the use of robust classifiers, which can deceive onlookers into believing that these explanations originate from the target classifier. The design of our ECED is illustrated in Figure~\ref{demo_framework}.

\subsection{Preliminary}
To explain the decision-making of the target classifier $f_{cl}$, our ECED framework takes an input image $\mathbf{\mathcal{I}}^\textrm{F}\in\mathbb{R}^{H_1\times W_1 \times 3}$ labeled as $\mathit{y}^\textrm{F}$ and determines the minimal alteration $\delta$ required to generate a counterfactual image $\mathbf{\mathcal{I}}^\textrm{CF}=\mathbf{\mathcal{I}}^\textrm{F} + \delta\in\mathbb{R}^{H_1\times W_1 \times 3}$ that achieves the desired label $\mathit{y}^\textrm{CF}$ with high probability when passed through $f_{cl}$. Firstly, our framework considers the influence of pixels $\mathbf{\mathcal{I}}^\textrm{F}_{i,j}$ on the label $\mathit{y}^\textrm{F}$ through the attention map $\mathbf{U}\in\mathbb{R}^{H_1\times W_1 \times 1}$ , $\mathbf{U}_{i,j}\in [0,1]$. Then we determine the foreground which is the critical region for editing, and the background which is the region to preserve in the image $\mathbf{\mathcal{I}}^\textrm{F}$ through a binary mask $\mathbf{\mathcal{M}}\in\mathbb{R}^{H_1\times W_1 \times 1}$, where $\mathbf{\mathcal{M}}_{i,j} = 1$ if $\mathbf{U}_{i,j} > \xi$ and vice versa. The counterfactual explanation ~$\mathbf{\mathcal{I}}^\textrm{CF}$ is achieved through an iterative procedure combining a blended diffusion process and adversarial attacks. Prior to transitioning to the counterfactual image generation phase, our method optimizes latent $\boldsymbol{z}^\textrm{bg}\in\mathbb{R}^{H_2\times W_2 \times 4}$ (we initialize: $\boldsymbol{z}^\textrm{bg}=\boldsymbol{z}^\textrm{F}$) and fine-tunes VAE Decoder $\theta_\mathcal{D}$ to minimize the deviation in the background of the reconstructed image from the original through the distance function $\mathcal{L}_{dis}$. At each iteration $t_2 = \overline{0,T_2}$ in the final stage, our framework performs diffusion processes on the initial latent $\boldsymbol{z}^\textrm{init}\in\mathbb{R}^{H_2\times W_2 \times 4}$ conditioned on $\mathbf{y}^\textrm{CF}$ represented by the text embedding $\mathbf{C}^\textrm{CF}\in\mathbb{R}^{O\times d}$, where $O$ is the number of tokens and $d$ is the token embedding dimension. This latent is the combination of the latent feature subsets $\boldsymbol{z}^\textrm{fg},\boldsymbol{z}^\textrm{bg*}\in\mathbb{R}^{H_2\times W_2 \times 4}$. Here, $\boldsymbol{z}^\textrm{fg}$ is updated through perturbation, and $\boldsymbol{z}^\textrm{bg*}$ has been optimized in the previous stage. Formally, the latent $\boldsymbol{z}^\textrm{init}$ is expressed as $\boldsymbol{z}^\textrm{init}=\boldsymbol{z}^\textrm{fg}\odot\mathbf{\mathcal{M}}'+\boldsymbol{z}^\textrm{bg*}\odot(\mathbf{1}-\mathbf{\mathcal{M}}')$, with the binary mask $\mathbf{\mathcal{M}}'\in\mathbb{R}^{H_2\times W_2 \times 1}$ downsampled from the original mask $\mathbf{\mathcal{M}}$. After generating each temporary counterfactual image, our method propagates the total Cross-Entropy loss $\mathcal{L}_{CE}$ back into $\boldsymbol{z}^\textrm{init}$ to make an adversarial gradient for input latent $\boldsymbol{z}^\textrm{init}$, step by step. Finally, we obtain the counterfactual explanation $\mathbf{\mathcal{I}}^\textrm{CF}=\mathcal{D}^*(\boldsymbol{z}_0)$ at last timestep $t_2 = T_2$.
\subsection{Identifying Key Regions}

To accurately classify objects in the family labels that can be confused, the observers should concentrate on features that are exclusive to one label and absent in the other labels. In this study, our main approach is to substitute the distinctive elements of the original label with the features corresponding to the desired label, ensuring these features are tightly interwoven with the remaining characteristics of the considered object. As illustrated in Figure~\ref{fig:ce-vce}, the vital regions corresponding to labels overlap, which can pose challenges in visualizing differences between the labels through visual explanations. However, the importance level for each pixel varies across corresponding labels within the intersecting region, i.e., $\mathbf{U}_{i,j} \neq \mathbf{U}'_{i,j}$ ($\mathbf{U}'_{i,j}$: target class-based attention map). Based on this observation, our method identifies the minimal and most crucial region in the image $\mathbf{\mathcal{I}}^\textrm{F}$ for the original label $\mathit{y}^\textrm{F}$, then considers this area as the foreground, which should have its features modified, whereas the remaining area is treated as the background and must remain unchanged in the counterfactual image $\mathbf{\mathcal{I}}^\textrm{CF}$. To accomplish this, we utilize visual explanation algorithms that rely on the CAM approach to extract the class activation map $\mathbf{U}$. The elements in the obtained map are subsequently compared with a predefined threshold $\xi\in[0,1)$ to obtain a binary mask $\mathbf{\mathcal{M}}$. Given the mask $\mathbf{\mathcal{M}}$, we establish the set $\mathcal{F} = \{(i,j) \mid \mathbf{\mathcal{M}}_{i,j}=1\}$, $\mathcal{B} = \{(i,j) \mid \mathbf{\mathcal{M}}_{i,j}=0\}$ as the collection of pixel positions $(i, j)$ that belong to the foreground and background set, respectively.

\subsection{Maintaining Background of the Input Image During Image Synthesis}
\label{sec:3.2}

As described in Section~\ref{sec:blended}, Avrahami~\textit{et al.}~\cite{avrahami2023blended}'s approach has fine-tuned the VAE Decoder to preserve the background after each generation process terminates. This operation results in a waste of computational resources overhead when our approach must manipulate the foreground in numerous iterations to obtain the image $\mathbf{\mathcal{I}}^\textrm{CF}$. To effectively adapt the Blended Latent Diffusion approach to this work, we propose optimizing certain components before transitioning to the counterfactual explanation generation phase. Specifically, our method optimizes background embedding and simultaneously fine-tunes the weights of the decoder $\boldsymbol{\theta}_{\mathcal{D}}$ to minimize the distance function $\mathcal{L}_{dis}$, as defined in Equation~\ref{eq:6}. The blended latent diffusion algorithm returns the reconstructed image by decoding the final latent $\boldsymbol{z}_0$. It is worth noting that this latent state retains a small amount of noise in the latent subspace that represents the background (see Supp for further discussion). Therefore, the goal of our method is to develop an optimization strategy to enhance the adaptability of the decoder at this state. Formally, the optimal latent $\boldsymbol{z}^\textrm{bg*}$ and the decoder parameters $\boldsymbol{\theta}^*_\mathcal{D}$ are those that minimize the following expression:
\vspace{-2mm}
\begin{align}
    \boldsymbol{z}^{\textrm{bg}^*}, \boldsymbol{\theta}^*_\mathcal{D} \in \operatorname*{arg\,min}\limits_{\boldsymbol{z}^{\textrm{bg}}, \boldsymbol{\theta}_\mathcal{D}} \lVert (\mathbf{1}-\mathbf{\mathcal{M}})\odot\left(\mathbf{\mathcal{I}}^\textrm{F}-\mathcal{D}(\boldsymbol{z}_0)\right) \rVert_p, \label{eq:6}
\end{align}
where $\boldsymbol{z}_0$ is obtained by adding noise to $\boldsymbol{z}^\textrm{init}$ according to Equation~\ref{eq:1} (we initialize $\boldsymbol{z}^\textrm{fg}=\boldsymbol{z}^\textrm{bg} =\boldsymbol{z}^\textrm{F}= \mathcal{E}(\mathbf{\mathcal{I}}^\textrm{F})$). According to Equation~\ref{eq:6}, we minimize the background content loss by using $L_p$ norms as the distance function.

\subsection{Generating Counterfactual Explanations}
\label{sec:3.4}
In this phase, our method generates counterfactual explanations by integrating the blended diffusion algorithm with adversarial attacks. This process is iteratively repeated, sequentially obtained from temporary hypothetical images $\mathbf{\mathcal{I}}^\textrm{(0)}, \mathbf{\mathcal{I}}^\textrm{(1)}, \mathbf{\mathcal{I}}^\textrm{(2)},\dots$, and ultimately produces counterfactual explanations $\mathbf{\mathcal{I}}^\textrm{CF}=\mathbf{\mathcal{I}}^{(T_2)}$. With each instance of acquiring an image $\mathbf{\mathcal{I}}^\textrm{(k)}$ ($k < T_2$), our framework computes the loss function $\mathbf{\mathcal{L}}_{CE}(f_{cl}(\mathbf{\mathcal{I}}^\textrm{(k)}), \mathit{y}^\textrm{CF})$, then makes gradient respect to $\boldsymbol{z}^\textrm{init}$ and updates it using variations of the gradient descent algorithm. Assuming that at each step $t_2 = k: \boldsymbol{\gamma}^\textrm{(k)}=\nabla_{\boldsymbol{z}^\textrm{init}}\mathcal{L}_{CE}(f_{cl}(\mathbf{\mathcal{I}}^\textrm{(k)}), \mathit{y}^\textrm{CF}) \in \mathbb{R}^{H_2\times W_2 \times 4}$ represents the fastest direction for achieving a highly realistic counterfactual image $\mathbf{\mathcal{I}}^\textrm{CF}$ for the desired label. Our goal is to preserve the background throughout the perturbation process. To this end, we implement an adversarial gradient pruning strategy by setting $\gamma_{i,j,s}^\textrm{(k)} = 0$ at positions $(i,j)$, where $\mathbf{\mathcal{M}}'_{i,j} = 0$ and $s$ denotes the channel order with $s = \overline{1,4}$. Given the change in gradient direction, our approach requires an increasing number of adversarial attack iterations to minimize the divergence between the distribution predicted by the target classifier and the distribution corresponding to the desired label $\mathit{y}^\textrm{CF}$. 

We delineate two primary concerns that must be addressed within the proposed methodology: i) whether the adversarial gradient pruning strategy guarantees convergence to the desired label, and ii) how to prune gradients effectively to induce sparse changes. Suppose $\xi^*$ is the optimal threshold to split foreground $\mathcal{F}^*$ and background $\mathcal{B}^*$. First, $c: \mathbb{R}^{H_2\times W_2\times 1} \rightarrow \mathbb{R}^{H_2\times W_2\times C}$ expands latent mask by concatenating this binary mask itself $C$ times along the channel dimension. Second, let $g: \mathbb{R}^{H_2\times W_2 \times C} \rightarrow \mathbb{R}^{H_2W_2C}$ map the space $H_2\times W_2 \times C$ to a vector of dimension $H_2W_2C$, where $H_2, W_2, C$ is the height, width, channel of latent space.

\begin{prop}
    Let $\boldsymbol{u}^* = g\left(\nabla_{\boldsymbol{z}^\textrm{init}}\mathcal{L}_{CE}\left(f_{cl}\left(\mathbf{\mathcal{I}}^\textrm{(k)}\right), \mathit{y}^\textrm{CF}\right)\right)$ is the optimal vector, and let $\boldsymbol{v}^* = g(c(\mathbf{\mathcal{M}}'))\odot\boldsymbol{u}^*$ be the vector representing the pruned adversarial gradient. Then
    \vspace{-3mm}
    \begin{align}
        0^\circ\leq\ \angle (\boldsymbol{u}^*,\boldsymbol{v}^*) <90^\circ,
    \end{align}
    where $\boldsymbol{u}^*= (u_{1,1,1},\dots, u_{H_2,W_2,1},\dots, u_{H_2,W_2,4})$, $u_{i,j,s}$ represents the element corresponding to $\boldsymbol{\gamma}^\textrm{(k)}_{i,j,s}, k = \overline{0, T_2}$. 
\end{prop}
 \vspace{-2.5mm}
\vspace{-1mm}

\vspace{-1mm}
\begin{prop}
    Suppose that $\{\xi_d\}^\infty_{d=1}$ is a sequence of thresholds satisfying:
    \begin{align}
        0\leq\xi_{d}<\xi_{d+1}< 1.
    \end{align}
    Then there exists a sequence of vectors $\left\{\boldsymbol{v}^{(d)}\right\}^\infty_{d=1}$ such that: 
    \begin{align}
        0^\circ\leq\angle \left(\boldsymbol{u}^*,\boldsymbol{v}^{(1)}\right)\leq\dots\leq\angle \left(\boldsymbol{u}^*,\boldsymbol{v}^{(\infty)}\right)<90^\circ. 
        \vspace{-1mm}
    \end{align}
\end{prop}

The detailed proofs of Propositions \ref{prop:proposition_1} and \ref{prop:proposition_2} are provided in the Supp. We propose a perspective on the trade-off between the ability to induce significant semantic changes and the convergence capability, based on determining an appropriate threshold $\xi$ during implementation to partition the image. More specifically, the chosen threshold $\xi_d$ must ensure that perturbations are applied over a sufficiently small region to prevent unnecessary modifications from inadvertently accelerating the convergence of the loss function $\mathcal{L}_{CE}$ to a local minimum. This allows the latent features representing the foreground to be significantly altered by gradients, as this process requires more iterations than usual. Therefore, this approach indirectly defines a sufficient region for recognizing label $\mathit{y}^\textrm{F}$, which is then replaced by semantically relevant details for label $\mathit{y}^\textrm{CF}$. Conversely, if the threshold $\xi_d$ is too large, it risks reducing the image space available for modification, potentially leading to a counterfactual image $\mathbf{\mathcal{I}}^\textrm{CF}$ that is not realistic and might not attain the desired label.

During the generation phase, our method executes local image perturbations by blending the latents $\boldsymbol{z}^\textrm{fg}, \boldsymbol{z}^\textrm{bg*}$. Through blending based on a binary mask $\mathbf{\mathcal{M}}'$ that is rescaled from the mask $\mathbf{\mathcal{M}}$ to match the dimensions of the latent space, $\boldsymbol{z}_t$ is the mixture of latent features representing the perturbed foreground and the fixed background. Thanks to the ability to leverage the context of Stable Diffusion, the counterfactual details are closely related to the preserved content (see Supp for further discussion). This study contributes a more lucid insight into the efficiency of aligning the latent space with the latent mask $\mathbf{\mathcal{M}'}$ that Avrahami~\textit{et al.}~\cite{avrahami2023blended} has not previously mentioned. This is because $\boldsymbol{z}^\textrm{init}$ captures high-level features of the depicted image, allowing this latent representation to convey more semantically meaningful concepts. This allows us to partially visualize the content of an image through the visualization of the latent space. The example can be found in Supp. 

\begin{figure*}
    \centering
    \includegraphics[width=0.9\linewidth]{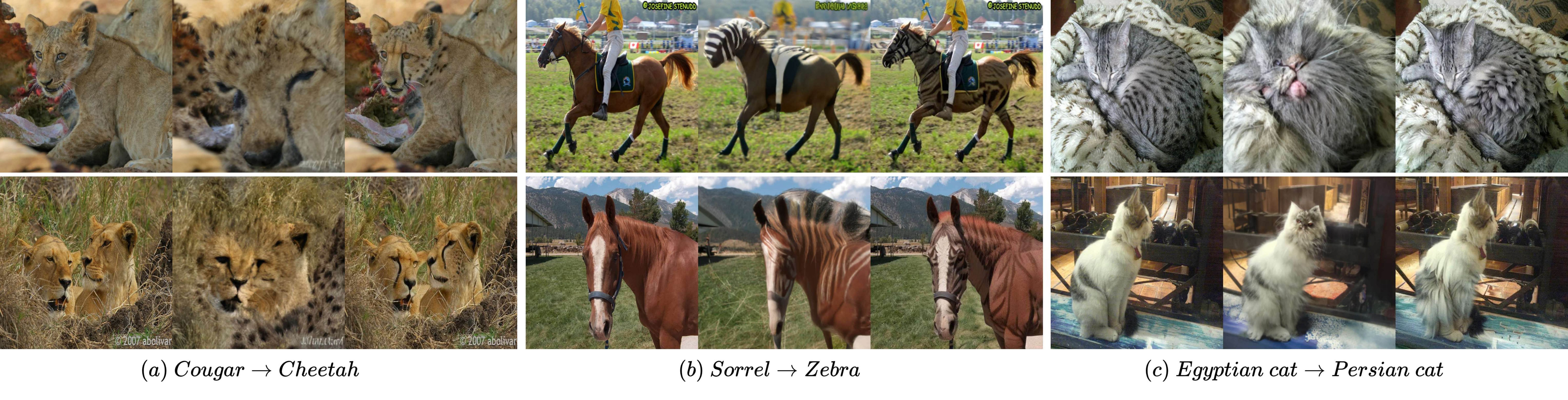}
    \vspace{-3mm}
    \caption{\textbf{Qualitative comparison on ImageNet.} From left to right: original image, counterfactual explanations for DVCE and ECED-r. }
    \label{fig:baseline_and_ECED}
    \vspace{-1.5mm}
\end{figure*}
\section{Experiments}
\label{sec:experiments}

\subsection{Datasets and Classification Models}
\label{subsec:datasets}
In this work, we evaluate our ECED model using two real-world datasets. Counterfactual explanations are computed on the 2000-image validation set of the CelebA-HQ \cite{lee2020maskgan} dataset with the VGG-16 classifier 
\begin{figure}
    \centering
    \includegraphics[width=0.7\linewidth]{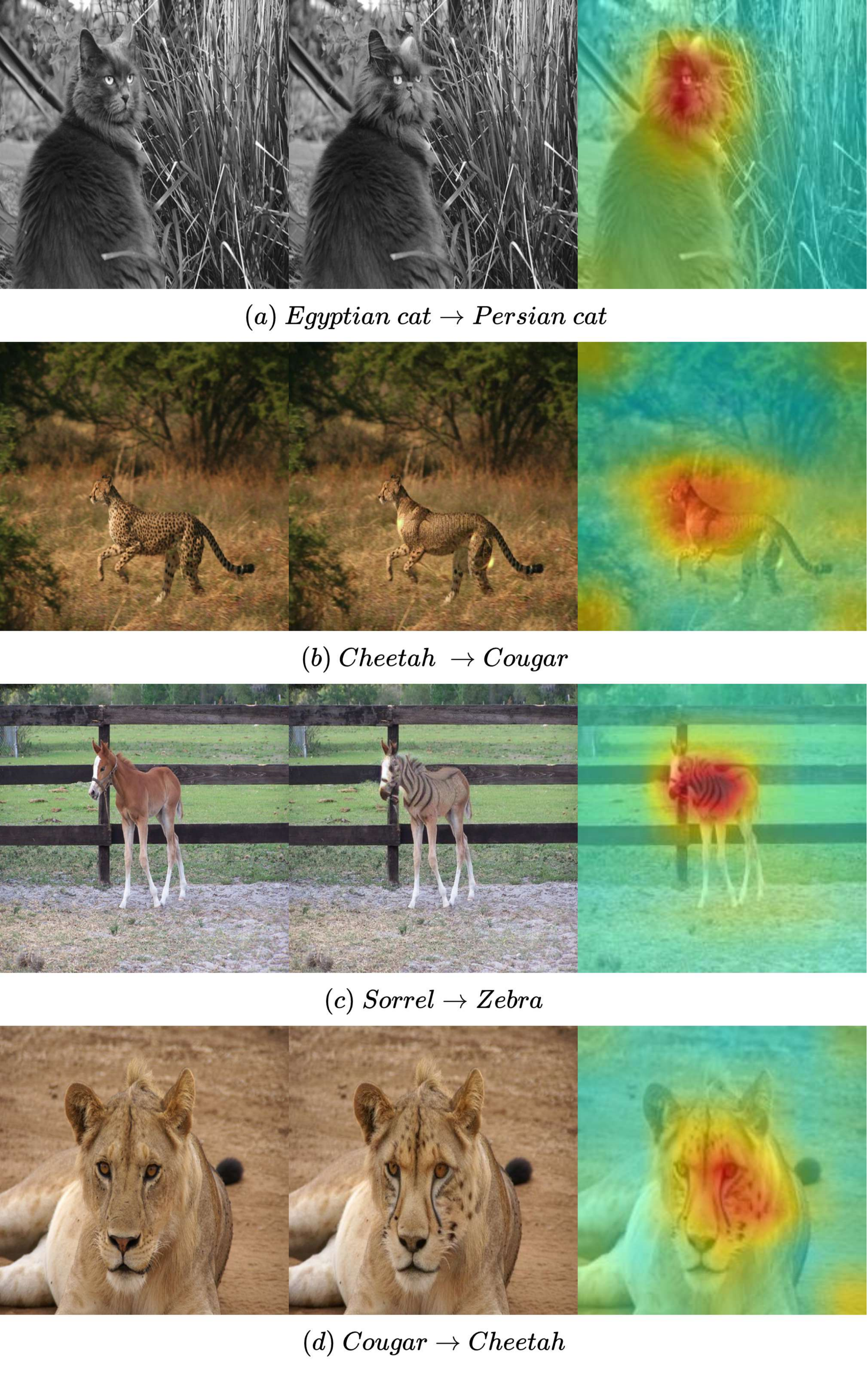}
    \vspace{-4mm}
    \caption{\textbf{Illustration of visual counterfactual explanations for ResNet-50.} From left to right: Original image, our counterfactual image, visual explanations for desired label.}
    \label{figure:4}
    \vspace{-3.5mm}
\end{figure}
\cite{simonyan2014very}, focusing on the `Smile' and `Age' attributes. Following Jeanneret~\textit{et al.}\cite{jeanneret2023adversarial}, we experimented with ECED on a small subset of ImageNet classes \cite{deng2009imagenet} consisting of 7800 images, using the VGG-16 and ResNet50 \cite{he2016deep}. All input images are rescaled to a resolution of 512 × 512 pixels, and the task is to generate counterfactual explanations (CEs) with labels opposite to those assigned to the samples. 

\subsection{Baselines and Evaluation Protocols}

\begin{table}[ht!]
\centering
\scriptsize
\caption{
\textbf{Quantitative Evaluation on ImageNet dataset.} We retrieved the ACE results from Jeanneret~\textit{et al.}~\cite{jeanneret2023adversarial}'s paper. ECED-v, ECED-r, ACE, and DVCE respectively generate CEs for VGG-16, ResNet50, ResNet50, and SwinT supported by MNR-RN50.
}
\label{table:2}
\setlength{\tabcolsep}{4pt} 
\renewcommand{\arraystretch}{1.2} 
\begin{tabular}{c|cccccccc} \toprule
        Method & $l_1$ $\downarrow$ & $l_{1.5}$ $\downarrow$ & $l_2$ $\downarrow$ & FID $\downarrow$  & sFID $\downarrow$  & S$^3$ $\uparrow$ & COUT $\uparrow$ & FR $\uparrow$\\ \midrule
        \multicolumn{9}{c}{\textbf{Zebra -- Sorrel}} \\\midrule
        ACE $\ell_1$ & - & - & - & 84.5 & 122.7 & \underline{0.9151} & -0.4462 & 47.0\\
        ACE $\ell_2$ & - & - & - & 67.7 & 98.4  & 0.9037 & -0.2525 & 81.0\\
        DVCE &  28373 & 622 & 99 & 124.9 & 154.9 & 0.3842 &  -0.0589 & 83.0\\\midrule
        ECED-v & \textbf{6733} & \textbf{192} & \textbf{39} & \textbf{33.0} & \textbf{45.21}	& \textbf{0.9501} & \textbf{0.7697}	& \underline{98.0}	\\
        ECED-r & \underline{16292} & \underline{437} & \underline{80} &  \underline{63.1} & \underline{97.8} & 0.8622 & \underline{0.6802} & \textbf{98.5} \\ \midrule
        
        \multicolumn{9}{c}{\textbf{Cheetah -- Cougar}}\\\midrule
        ACE $\ell_1$ & - & - & - & 70.2 &100.5 & 0.9085 & 0.0173   & 77.0 \\
        ACE $\ell_2$ & - & - & - & 74.1 & 102.5 &  0.8785 & 0.1203   & 95.0 \\
        DVCE   & 28757 & 615 & 96  & 122.3 & 131.7	& 0.2992 & 0.6520 & \textbf{100.0}	\\\hline
        ECED-v &   \textbf{7761}   & \textbf{208}  & \textbf{40} & \textbf{23.3}   & \textbf{37.3}        & \underline{0.9486} & \textbf{0.8310} & \underline{99.5} \\ 
        ECED-r  & \underline{8917} & \underline{249} & \underline{48}  & \underline{47.0} & \underline{76.6} & \textbf{0.9598} & \underline{0.8230} & \textbf{100.0} \\\midrule
        
        \multicolumn{9}{c}{\textbf{Egyptian Cat -- Persian Cat}} \\\midrule
        ACE $\ell_1$ & - & - & - & 93.6 & 156.7 &  0.8467 &  0.2491 & 85.0 \\
        ACE $\ell_2$ & - & - & - & 107.3 & 160.4 & 0.7810 &  0.3430  & \underline{97.0} \\
        DVCE & 19857 & 451 & 74  & 163.2 & 176.3	& 0.3589 & 0.7328 &	\textbf{100.0}	    \\ \hline

        ECED-v & \textbf{6765} & \textbf{172} & \textbf{32}  & \textbf{27.4} & \textbf{53.5} & \textbf{0.9806}& \textbf{0.8546} & \textbf{100.0}	        \\
        ECED-r  & \underline{7776} & \underline{203} &  \underline{37}  & \underline{44.3} & \underline{71.2} & \underline{0.9378} & \underline{0.8102} & \textbf{100.0}	     \\
        \bottomrule
    \end{tabular}
    \vspace{-4mm}
\end{table}
\begin{figure*}
    \centering
    \includegraphics[width=0.7\linewidth]{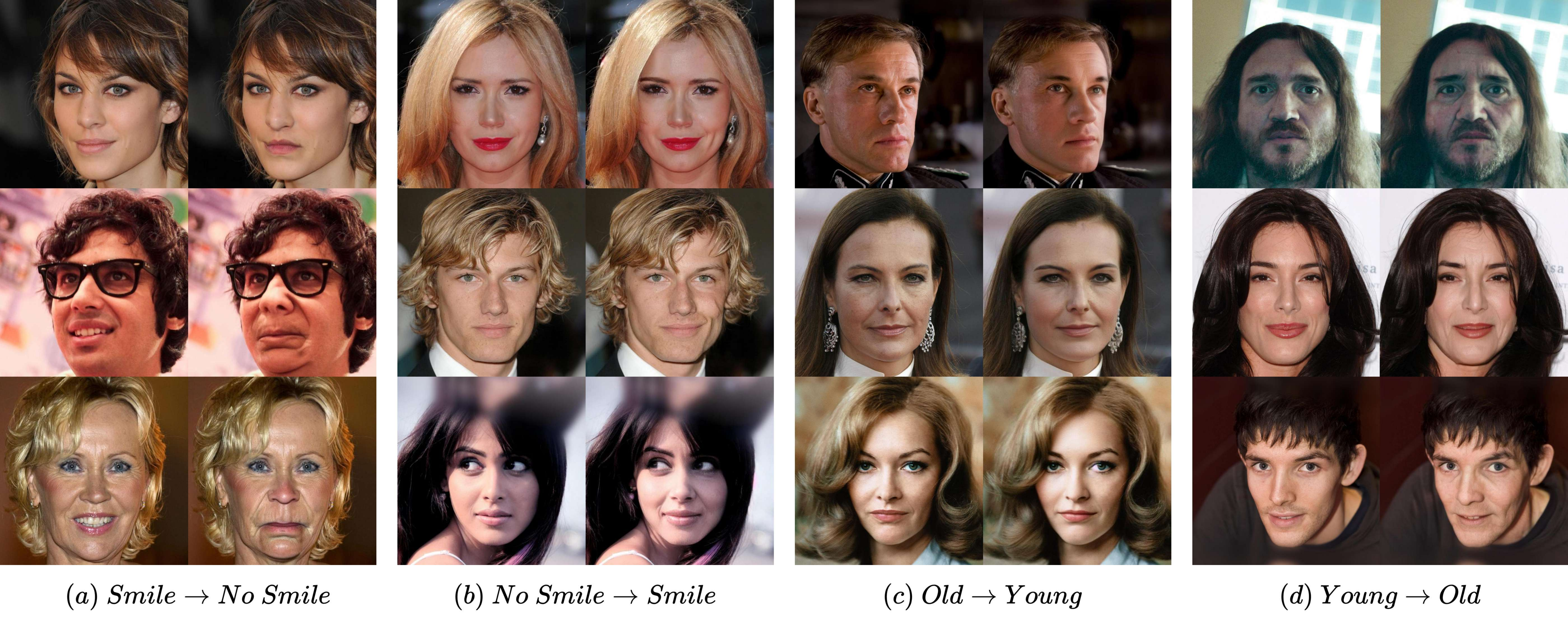}
    \vspace{-3mm}
    \caption{\textbf{Qualitative results on CelebA-HQ with VGG-16.} Left to right: original image, counterfactual image generated by ECED.}
    \label{fig:CelebAHQ_ex}
\end{figure*}
\begin{table*}[t]
    \centering
    \scriptsize
    \setlength{\tabcolsep}{6pt} 
    \caption{\textbf{Quantitative Evaluation on CelebA-HQ dataset.}We retrieved the previous results from the ACE paper. In various assessment protocols, our ECED outperforms most recent prominent works.}
    \begin{tabular}{c|ccccccc|ccccccc} \toprule
        \multicolumn{1}{c}{} & \multicolumn{7}{|c|}{\textbf{Smile}} & \multicolumn{7}{c}{\textbf{Age}}\\\midrule
        Method        & FID $\downarrow$  & sFID $\downarrow$ & FVA $\uparrow$  & FS $\uparrow$ & MNAC $\downarrow$ & CD $\downarrow$   & COUT $\uparrow$ & FID $\downarrow$  & sFID $\downarrow$ & FVA $\uparrow$  & FS $\uparrow$ & MNAC $\downarrow$ & CD $\downarrow$   & COUT $\uparrow$ \\ \midrule
        DiVE          &107.0 & -    & 35.7 & -  & 7.41  & -     & -    &107.5 & -    & 32.3 & -  & 6.76 & -    & - \\ 
        STEEX         & 21.9 & -    & \underline{97.6} & -  & 5.27  & -     & -    & 26.8 & -    & 96.0 & -  & 5.63 & -    & - \\
        DiME          & 18.1 & 27.7 & 96.7 & 0.6729 & 2.63  & \textbf{1.82}  & \underline{0.6495} & 18.7 & 27.8 & 95.0 & 0.6597 & 2.10 & \underline{4.29} & \underline{0.5615} \\
        ACE $\ell_1$  & \textbf{3.21} & \textbf{20.2} & \textbf{100.0} & \underline{0.8941} & \underline{1.56}  & 2.61  & 0.5496 & \textbf{5.31} & \textbf{21.7} & \underline{99.6} & \underline{0.8085} & \underline{1.53} & 5.4  & 0.3984 \\
        ACE $\ell_2$  & \underline{6.93} & \underline{22.0} & \textbf{100.0} & 0.8440 & 1.87  & \underline{2.21}  & 0.5946 & 16.4 & \underline{28.2} & \underline{99.6} & 0.7743 & 1.92 & \textbf{4.21} & 0.5303 \\\midrule
        ECED & 7.56 & 24.9 & \textbf{100.0} & \textbf{0.9021} & \textbf{1.32}  &   5.08 & \textbf{0.8421} & \underline{12.4} & 31.6 & \textbf{100.0} & \textbf{0.9563} & \textbf{0.84} & 4.93 & \textbf{0.7442} \\
        \bottomrule
    \end{tabular}
    \vspace{-2.5mm}
    \label{tab:1}
\end{table*}
\textbf{State-of-the-art models.} We compare our method to two notable studies: DVCE \cite{augustin2022diffusion} and ACE \cite{avrahami2023blended}. Both of these studies are capable of producing counterfactual images that closely correspond to human perception and achieve high accuracy in identifying the desired label. We directly refer to the ACE study's results, which evaluated the CEs using a ResNet50 for ImageNet and a DenseNet121 \cite{huang2017densely} for CelebA-HQ. For DVCE, we conducted experiments on subsets of ImageNet, as mentioned in Section~\ref{subsec:datasets}. We followed the guidelines and utilized the checkpoints provided by Augustin~\textit{et al.} to generate CEs for the Swin-TF model \cite{liu2021swin}, supported by the pretrained multiple-norm robust ResNet50 (MNR-RN50) \cite{boreiko2022sparse}, with a degree of cone projection set to 30. To obtain a comprehensive understanding of these hyperparameters, please refer to the DVCE work \cite{augustin2022diffusion}.

\textbf{Assessment criteria for Quantitative Evaluation.} We compare the results of the ECED method with previous works based on critical evaluation protocols. \textit{Sparsity:} $l_p$ norm ($p=\{1,1.5,2\}$) and $S^3$ \cite{chen2021exploring} measure the similarity between the original images and their corresponding counterfactuals. For the CelebA-HQ dataset, Face Verification Accuracy (\underline{FVA}) and Face Similarity (\underline{FS}) \cite{jeanneret2023adversarial} quantify the ability to preserve the face-based identity of counterfactual images. Mean Number of Attribute Changes (\underline{MNAC}) \cite{rodriguez2021beyond} and Correlation Difference (\underline{CD}) \cite{jeanneret2022diffusion} quantify the minimum number of traits that are altered between the original and counterfactual image. Counterfactual Transition (\underline{COUT}) \cite{khorram2022cycle} measures the sparse changes required in the original image to achieve the desired label. \textit{Realism:} Fréchet Inception Distance (\underline{FID}) \cite{heusel2017gans} and split FID (\underline{sFID}) \cite{jeanneret2023adversarial} assess whether the distribution of the counterfactual image set closely aligns with that of the original set. It should be noted that the sFID value varies depending on how the subsets are divided. \textit{Validity:} Flip Ratio (\underline{FR}) evaluates the confidence level of the CEs in achieving the desired label. The evaluation protocols are detailed in Supp.

\subsection{Implementation Details}
\label{sec:implement}
We used the \texttt{pruned v1-5 checkpoint} of Stable Diffusion (mini version), which was trained on the large real-world datasets LAION-5B \cite{schuhmann2022laion}, to generate counterfactual images for both the ImageNet and CelebA-HQ datasets. For the CelebA-HQ dataset, we fine-tuned Stable Diffusion to adapt to prompts aligned with the desired conditions (details can be found in Supp). The denoising process was carried out within the scheduler $\tau = 5$ out of 50 steps to ensure the generated images closely resembled the original images. To provide visual explanations for the initial labels, the ScoreCAM method \cite{wang2020score} was used. Activation maps were extracted from specific convolutional layers, close to the fully-connected layers, by setting $layer\_name$ to `features\_29' for VGG-16. For ResNet50, we set $layer\_name$ to `layer4\_bottleneck1' (please refer to the guidelines mentioned in ScoreCAM \cite{wang2020score} for a detailed understanding of the configuration). We set $\xi = 0.5$ to create a sufficiently adequate foreground region to balance the trade-off between semantically meaningful changes and sparse changes. The $l_{1.5}$ loss function was utilized to optimize the VAE Decoder and the latent space representing the foreground over 500 steps, with the learning rate of $3.10^{-4}$. Unlike the ACE framework, we perform adversarial attacks on the latent space with a value range of $\mathbb{R}^4$ for each feature instead of the pixel space with a range of $[0,255]^3$ for each pixel. Consequently, rather than scaling to a specific value range, we need to control the magnitude of gradient added to the latent at each iteration. To address this issue, we employ the Adam optimizer \cite{kingma2014adam}, taking advantage of its flexibility in hyperparameter tuning. We set $\beta_1 = 0.9, \beta_2 = 0.999$, with a learning rate of $7.10^{-2}$. All experiments were conducted on an A100 SXM4 80GB GPU. Code will be available at \url{https://github.com/tungluuai/ECED}.
\subsection{Experimental Results}
\label{subsec: Experimental Results}
\textbf{Quantitative Evaluation.} The experimental results on the ImageNet dataset demonstrate that our ECED outperforms the compared baselines across all evaluation criteria, as shown in Table~\ref{table:2}. By observing COUT scores, the results of ACE and DVCE indicate that CEs need to undergo significant changes to achieve high validity on the target classifiers. Our approach effectively addresses the trade-off between sparse changes and high validity by perturbing the minimal and essential regions of the original image. Furthermore, the distribution of counterfactual images generated by these methods significantly deviates from the ImageNet dataset distribution, whereas our approach substantially mitigates this limitation. Visual counterfactual explanations must closely resemble real-world data, which poses a significant challenge for CE methods when applied to the ImageNet dataset. Our ECED addresses this challenge by leveraging the context latent and denoising on the latent subspace representing the foreground, whereas ACE and DVCE approaches denoise across the entire pixel space. For the CelebA-HQ dataset, Table~\ref{tab:1} shows that the ACE results have slightly better FID and sFID scores compared to ours. We attribute the primary reason to the suboptimal fine-tuning strategy of Stable Diffusion, whereas Jeanneret~\textit{et al.}'s ACE leveraged the diffusion model \cite{dhariwal2021diffusion} pre-trained on the CelebA-HQ dataset. Regarding the CD metric, our results only surpass ACE $l_1$ for the `Age' attribute. The CD values were computed based on a pretrained ResNet50, meanwhile we generated the CEs using a VGG-16 classifier. For other criteria related to identity preservation and generating sparse changes, our method indicates more favorable results than all other methods. 

\textbf{Qualitative Evaluation.} As depicted in Figure~\ref{fig:baseline_and_ECED}, our hypothetical images satisfy the desired properties of Counterfactual Explanations. For instance, from the original class `Cougar' to the desired class `Cheetah', the cougar cub's face was modified with black stripes and spots, while the surrounding grass and rocks remained unchanged. In contrast, the counterfactual images generated by the DVCE method do not focus on perturbing the object, which may be attributed to the fact that DVCE performs diffusion across the entire image rather than on a specific region. Figure~\ref{fig:CelebAHQ_ex} shows that the CEs are predicted by the target classifier as the desired label with high probability. These hypothetical images align intuitively with human understanding, such as smoothing the skin to change from `Old' to `Young' or adding wrinkles to change from `Young' to `Old'. More results can be viewed in Supp.
\begin{table}[]
    \centering
    \scriptsize
    \caption{\textbf{Trade-off between Validity and Sparsity:} Comparison of two threshold scenarios $\xi = 0.5$ and $\xi = 0.7$. We conducted the study over 100 random samples.}
    \begin{tabular}{c|cc}
    \toprule
                                                                            Threshold & COUT $\uparrow$ & FR $\uparrow$ \\ 
    \midrule
    CEs with $\xi = 0.5$ & 0.79                                & \textbf{100.0}                     \\ 
    \midrule
    CEs with $\xi = 0.7$ & \textbf{0.84}                       & 95.0                              \\
    \bottomrule
    \end{tabular}
    \vspace{-5mm}
    \label{trade-off}
\end{table}

\textbf{Semantic Changes.} After making semantically meaningful changes, we aim to verify whether the target classifier relies on these details to alter its decision-making. Using the ScoreCAM algorithm, we conducted experiments on several counterfactual images and visualized visual explanations for the desired labels. Figure~\ref{figure:4} demonstrates that the class activation maps highlight the most distinguishing features between the counterfactual and the original images.

\textbf{Background Preservation.} We examined the distance loss analysis for a sample of 100 randomly selected images. The findings shown in Table~\ref{table:3} illustrate the efficacy of our method since the pixel values in the background area show little variations in comparison to the original pixel values. Hence, the background material remains essentially unaltered when observed visually.

\begin{figure}
    \centering
    \includegraphics[width=0.7\linewidth]{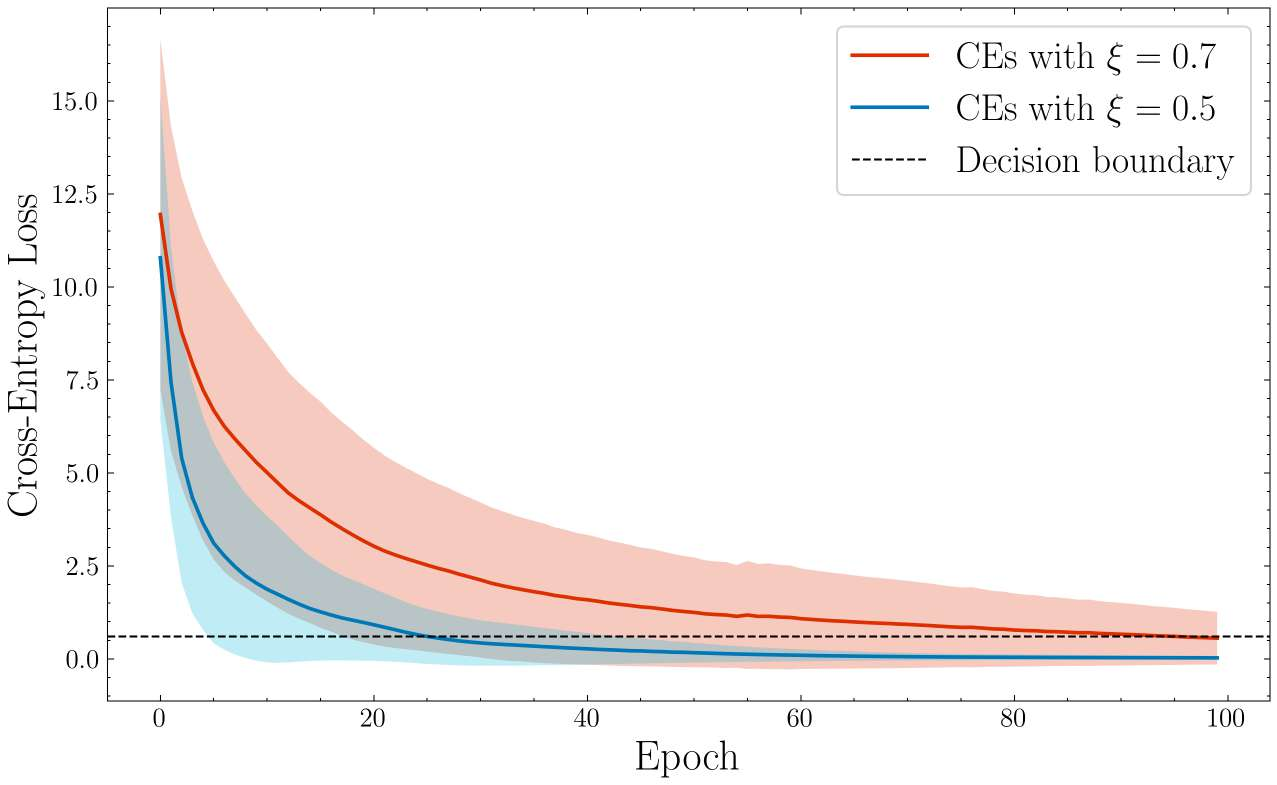}
    \vspace{-2mm}
    \caption{\textbf{Convergence Towards Desired Labels:} Comparison of two threshold scenarios $\xi = 0.5$ and $\xi = 0.7$. For each threshold $\xi$, the curves and shaded areas represent the mean cross-entropy loss and the $95\%$ confidence interval of the mean values over 100 epochs. We conducted an analysis over 100 random images.}
    \label{convergence_comparison}
    \vspace{-2mm}
\end{figure}

\textbf{Convergence and Trade-off.} The theoretical analyses presented in Section~\ref{sec:3.4} are clearly illustrated by the experimental results shown in Table~\ref{trade-off}. Specifically, with a larger threshold  $\xi$, we obtain visual CEs with sparser changes; however, there exist counterfactual images that are not predicted as the desired labels. This issue is reflected in the convergence visualized in Figure~\ref{convergence_comparison}. With a smaller threshold $\xi$, all CEs achieve the desired label after approximately 40 epochs and exhibit high validity after around 60 epochs. Meanwhile, with $\xi = 0.7$, on average, CEs only achieve the desired label in the last few epochs, and some do not achieve it based on upper-bound observations.

\textbf{Limitation.} The main constraint of our approach is the temporal requirements for generating a counterfactual explanation. We report that our approach required roughly 150 seconds for a counterfactual image, whereas the ACE and DVCE approaches took 184 seconds and 52 seconds, respectively. Another drawback of our method is the requirement to manually adjust the foreground-background separation hyperparameter to enhance the optimization or generate realistic counterfactual images.
\begin{table}[]
    \centering
    \scriptsize
    \caption{\textbf{Background Content Loss.} We calculated the Mean Squared Error (MSE) loss for each pixel at the points $\mathcal{P}_1$: preserved image after optimization in the second phase, $\mathcal{P}_2$: visual counterfactual explanation $\mathbf{\mathcal{I}}^\textrm{CF}$, and averaged the MSE loss over 100 iterations with $95\%$ confidence interval.}
    \begin{tabular}{ccc}
        \toprule
        \multicolumn{3}{c}{\textbf{Background Content Loss}}\\
        \midrule
        $\mathcal{P}_1$ & $\mathcal{P}_2$ & Avg. \\
        \midrule
        0.0050 $\pm$ 0.0009 & 0.0081 $\pm$ 0.0014 & 0.0077 $\pm$ 0.0012\\
        \bottomrule
    \end{tabular}
    \vspace{-5mm}
    \label{table:3}
\end{table}
\section{Conclusion}
\label{sec:conclusion}
This work presents a technique for generating counterfactual explanations for the target classifier without relying on an adversarially robust model. ECED utilizes a class activation map to identify the critical regions, and subsequently edits these regions by integrating the latent diffusion model with adversarial gradient pruning-based attacks. Our method effectively addresses the trade-off between criteria in quantitative evaluation and shows that the counterfactual explanations possess a high level of semantic value.
\paragraph{Acknowledgements.}
This research is funded by Vietnam National Foundation for Science and Technology Development (NAFOSTED) under grant number 102.05-2023.44. And this research used the GPUs provided by the Intelligent Systems Lab at the Faculty of Information Technology, University of Science, VNU-HCM. 
{\small
\bibliographystyle{ieee_fullname}
\bibliography{egbib}

\begin{thebibliography}{10}\itemsep=-1pt

\bibitem{augustin2022diffusion}
Maximilian Augustin, Valentyn Boreiko, Francesco Croce, and Matthias Hein.
\newblock Diffusion visual counterfactual explanations.
\newblock {\em Advances in Neural Information Processing Systems}, 35:364--377, 2022.

\bibitem{avrahami2023blended}
Omri Avrahami, Ohad Fried, and Dani Lischinski.
\newblock Blended latent diffusion.
\newblock {\em ACM Transactions on Graphics (TOG)}, 42(4):1--11, 2023.

\bibitem{boreiko2022sparse}
Valentyn Boreiko, Maximilian Augustin, Francesco Croce, Philipp Berens, and Matthias Hein.
\newblock Sparse visual counterfactual explanations in image space.
\newblock In {\em DAGM German Conference on Pattern Recognition}, pages 133--148. Springer, 2022.

\bibitem{chattopadhay2018grad}
Aditya Chattopadhay, Anirban Sarkar, Prantik Howlader, and Vineeth~N Balasubramanian.
\newblock Grad-cam++: Generalized gradient-based visual explanations for deep convolutional networks.
\newblock In {\em 2018 IEEE winter conference on applications of computer vision (WACV)}, pages 839--847. IEEE, 2018.

\bibitem{chen2021exploring}
Xinlei Chen and Kaiming He.
\newblock Exploring simple siamese representation learning.
\newblock In {\em Proceedings of the IEEE/CVF conference on computer vision and pattern recognition}, pages 15750--15758, 2021.

\bibitem{deng2009imagenet}
Jia Deng, Wei Dong, Richard Socher, Li-Jia Li, Kai Li, and Li Fei-Fei.
\newblock Imagenet: A large-scale hierarchical image database.
\newblock In {\em 2009 IEEE conference on computer vision and pattern recognition}, pages 248--255. Ieee, 2009.

\bibitem{dhariwal2021diffusion}
Prafulla Dhariwal and Alexander Nichol.
\newblock Diffusion models beat gans on image synthesis.
\newblock {\em Advances in neural information processing systems}, 34:8780--8794, 2021.

\bibitem{goodfellow2020generative}
Ian Goodfellow, Jean Pouget-Abadie, Mehdi Mirza, Bing Xu, David Warde-Farley, Sherjil Ozair, Aaron Courville, and Yoshua Bengio.
\newblock Generative adversarial networks.
\newblock {\em Communications of the ACM}, 63(11):139--144, 2020.

\bibitem{he2016deep}
Kaiming He, Xiangyu Zhang, Shaoqing Ren, and Jian Sun.
\newblock Deep residual learning for image recognition.
\newblock In {\em Proceedings of the IEEE conference on computer vision and pattern recognition}, pages 770--778, 2016.

\bibitem{hertz2022prompt}
Amir Hertz, Ron Mokady, Jay Tenenbaum, Kfir Aberman, Yael Pritch, and Daniel Cohen-Or.
\newblock Prompt-to-prompt image editing with cross attention control.
\newblock {\em arXiv preprint arXiv:2208.01626}, 2022.

\bibitem{heusel2017gans}
Martin Heusel, Hubert Ramsauer, Thomas Unterthiner, Bernhard Nessler, and Sepp Hochreiter.
\newblock Gans trained by a two time-scale update rule converge to a local nash equilibrium.
\newblock {\em Advances in neural information processing systems}, 30, 2017.

\bibitem{ho2020denoising}
Jonathan Ho, Ajay Jain, and Pieter Abbeel.
\newblock Denoising diffusion probabilistic models.
\newblock {\em Advances in neural information processing systems}, 33:6840--6851, 2020.

\bibitem{huang2017densely}
Gao Huang, Zhuang Liu, Laurens Van Der~Maaten, and Kilian~Q Weinberger.
\newblock Densely connected convolutional networks.
\newblock In {\em Proceedings of the IEEE conference on computer vision and pattern recognition}, pages 4700--4708, 2017.

\bibitem{jeanneret2022diffusion}
Guillaume Jeanneret, Lo{\"\i}c Simon, and Fr{\'e}d{\'e}ric Jurie.
\newblock Diffusion models for counterfactual explanations.
\newblock In {\em Proceedings of the Asian Conference on Computer Vision}, pages 858--876, 2022.

\bibitem{jeanneret2023adversarial}
Guillaume Jeanneret, Lo{\"\i}c Simon, and Fr{\'e}d{\'e}ric Jurie.
\newblock Adversarial counterfactual visual explanations.
\newblock In {\em Proceedings of the IEEE/CVF Conference on Computer Vision and Pattern Recognition}, pages 16425--16435, 2023.

\bibitem{jiang2021layercam}
Peng-Tao Jiang, Chang-Bin Zhang, Qibin Hou, Ming-Ming Cheng, and Yunchao Wei.
\newblock Layercam: Exploring hierarchical class activation maps for localization.
\newblock {\em IEEE Transactions on Image Processing}, 30:5875--5888, 2021.

\bibitem{khorram2022cycle}
Saeed Khorram and Li Fuxin.
\newblock Cycle-consistent counterfactuals by latent transformations.
\newblock In {\em Proceedings of the IEEE/CVF Conference on Computer Vision and Pattern Recognition}, pages 10203--10212, 2022.

\bibitem{kingma2014adam}
Diederik~P Kingma and Jimmy Ba.
\newblock Adam: A method for stochastic optimization.
\newblock {\em arXiv preprint arXiv:1412.6980}, 2014.

\bibitem{kingma2013auto}
Diederik~P Kingma and Max Welling.
\newblock Auto-encoding variational bayes.
\newblock {\em arXiv preprint arXiv:1312.6114}, 2013.

\bibitem{lee2020maskgan}
Cheng-Han Lee, Ziwei Liu, Lingyun Wu, and Ping Luo.
\newblock Maskgan: Towards diverse and interactive facial image manipulation.
\newblock In {\em Proceedings of the IEEE/CVF conference on computer vision and pattern recognition}, pages 5549--5558, 2020.

\bibitem{liu2024towards}
Bingyan Liu, Chengyu Wang, Tingfeng Cao, Kui Jia, and Jun Huang.
\newblock Towards understanding cross and self-attention in stable diffusion for text-guided image editing.
\newblock In {\em Proceedings of the IEEE/CVF Conference on Computer Vision and Pattern Recognition}, pages 7817--7826, 2024.

\bibitem{liu2021swin}
Ze Liu, Yutong Lin, Yue Cao, Han Hu, Yixuan Wei, Zheng Zhang, Stephen Lin, and Baining Guo.
\newblock Swin transformer: Hierarchical vision transformer using shifted windows.
\newblock In {\em Proceedings of the IEEE/CVF international conference on computer vision}, pages 10012--10022, 2021.

\bibitem{loshchilov2017decoupled}
I Loshchilov.
\newblock Decoupled weight decay regularization.
\newblock {\em arXiv preprint arXiv:1711.05101}, 2017.

\bibitem{lugmayr2022repaint}
Andreas Lugmayr, Martin Danelljan, Andres Romero, Fisher Yu, Radu Timofte, and Luc Van~Gool.
\newblock Repaint: Inpainting using denoising diffusion probabilistic models.
\newblock In {\em Proceedings of the IEEE/CVF conference on computer vision and pattern recognition}, pages 11461--11471, 2022.

\bibitem{madry2017towards}
Aleksander Madry, Aleksandar Makelov, Ludwig Schmidt, Dimitris Tsipras, and Adrian Vladu.
\newblock Towards deep learning models resistant to adversarial attacks.
\newblock {\em arXiv preprint arXiv:1706.06083}, 2017.

\bibitem{moosavi2016deepfool}
Seyed-Mohsen Moosavi-Dezfooli, Alhussein Fawzi, and Pascal Frossard.
\newblock Deepfool: a simple and accurate method to fool deep neural networks.
\newblock In {\em Proceedings of the IEEE conference on computer vision and pattern recognition}, pages 2574--2582, 2016.

\bibitem{naidu2020cam}
Rakshit Naidu, Ankita Ghosh, Yash Maurya, Soumya~Snigdha Kundu, et~al.
\newblock Is-cam: Integrated score-cam for axiomatic-based explanations.
\newblock {\em arXiv preprint arXiv:2010.03023}, 2020.

\bibitem{papernot2016transferability}
Nicolas Papernot, Patrick McDaniel, and Ian Goodfellow.
\newblock Transferability in machine learning: from phenomena to black-box attacks using adversarial samples.
\newblock {\em arXiv preprint arXiv:1605.07277}, 2016.

\bibitem{papernot2016limitations}
Nicolas Papernot, Patrick McDaniel, Somesh Jha, Matt Fredrikson, Z~Berkay Celik, and Ananthram Swami.
\newblock The limitations of deep learning in adversarial settings.
\newblock In {\em 2016 IEEE European symposium on security and privacy (EuroS\&P)}, pages 372--387. IEEE, 2016.

\bibitem{radford2021learning}
Alec Radford, Jong~Wook Kim, Chris Hallacy, Aditya Ramesh, Gabriel Goh, Sandhini Agarwal, Girish Sastry, Amanda Askell, Pamela Mishkin, Jack Clark, et~al.
\newblock Learning transferable visual models from natural language supervision.
\newblock In {\em International conference on machine learning}, pages 8748--8763. PMLR, 2021.

\bibitem{rodriguez2021beyond}
Pau Rodriguez, Massimo Caccia, Alexandre Lacoste, Lee Zamparo, Issam Laradji, Laurent Charlin, and David Vazquez.
\newblock Beyond trivial counterfactual explanations with diverse valuable explanations.
\newblock In {\em Proceedings of the IEEE/CVF International Conference on Computer Vision}, pages 1056--1065, 2021.

\bibitem{ronneberger2015u}
Olaf Ronneberger, Philipp Fischer, and Thomas Brox.
\newblock U-net: Convolutional networks for biomedical image segmentation.
\newblock In {\em Medical image computing and computer-assisted intervention--MICCAI 2015: 18th international conference, Munich, Germany, October 5-9, 2015, proceedings, part III 18}, pages 234--241. Springer, 2015.

\bibitem{samangouei2018explaingan}
Pouya Samangouei, Ardavan Saeedi, Liam Nakagawa, and Nathan Silberman.
\newblock Explaingan: Model explanation via decision boundary crossing transformations.
\newblock In {\em Proceedings of the European Conference on Computer Vision (ECCV)}, pages 666--681, 2018.

\bibitem{schuhmann2022laion}
Christoph Schuhmann, Romain Beaumont, Richard Vencu, Cade Gordon, Ross Wightman, Mehdi Cherti, Theo Coombes, Aarush Katta, Clayton Mullis, Mitchell Wortsman, et~al.
\newblock Laion-5b: An open large-scale dataset for training next generation image-text models.
\newblock {\em Advances in Neural Information Processing Systems}, 35:25278--25294, 2022.

\bibitem{selvaraju2020grad}
Ramprasaath~R Selvaraju, Michael Cogswell, Abhishek Das, Ramakrishna Vedantam, Devi Parikh, and Dhruv Batra.
\newblock Grad-cam: visual explanations from deep networks via gradient-based localization.
\newblock {\em International journal of computer vision}, 128:336--359, 2020.

\bibitem{selvaraju2016grad}
Ramprasaath~R Selvaraju, Abhishek Das, Ramakrishna Vedantam, Michael Cogswell, Devi Parikh, and Dhruv Batra.
\newblock Grad-cam: Why did you say that?
\newblock {\em arXiv preprint arXiv:1611.07450}, 2016.

\bibitem{simonyan2014very}
Karen Simonyan and Andrew Zisserman.
\newblock Very deep convolutional networks for large-scale image recognition.
\newblock {\em arXiv preprint arXiv:1409.1556}, 2014.

\bibitem{sohl2015deep}
Jascha Sohl-Dickstein, Eric Weiss, Niru Maheswaranathan, and Surya Ganguli.
\newblock Deep unsupervised learning using nonequilibrium thermodynamics.
\newblock In {\em International conference on machine learning}, pages 2256--2265. PMLR, 2015.

\bibitem{song2020denoising}
Jiaming Song, Chenlin Meng, and Stefano Ermon.
\newblock Denoising diffusion implicit models.
\newblock {\em arXiv preprint arXiv:2010.02502}, 2020.

\bibitem{szegedy2016rethinking}
Christian Szegedy, Vincent Vanhoucke, Sergey Ioffe, Jon Shlens, and Zbigniew Wojna.
\newblock Rethinking the inception architecture for computer vision.
\newblock In {\em Proceedings of the IEEE conference on computer vision and pattern recognition}, pages 2818--2826, 2016.

\bibitem{wang2020ss}
Haofan Wang, Rakshit Naidu, Joy Michael, and Soumya~Snigdha Kundu.
\newblock Ss-cam: Smoothed score-cam for sharper visual feature localization.
\newblock {\em arXiv preprint arXiv:2006.14255}, 2020.

\bibitem{wang2020score}
Haofan Wang, Zifan Wang, Mengnan Du, Fan Yang, Zijian Zhang, Sirui Ding, Piotr Mardziel, and Xia Hu.
\newblock Score-cam: Score-weighted visual explanations for convolutional neural networks.
\newblock In {\em Proceedings of the IEEE/CVF conference on computer vision and pattern recognition workshops}, pages 24--25, 2020.

\bibitem{wang2018high}
Ting-Chun Wang, Ming-Yu Liu, Jun-Yan Zhu, Andrew Tao, Jan Kautz, and Bryan Catanzaro.
\newblock High-resolution image synthesis and semantic manipulation with conditional gans.
\newblock In {\em Proceedings of the IEEE conference on computer vision and pattern recognition}, pages 8798--8807, 2018.

\bibitem{zhou2016learning}
Bolei Zhou, Aditya Khosla, Agata Lapedriza, Aude Oliva, and Antonio Torralba.
\newblock Learning deep features for discriminative localization.
\newblock In {\em Proceedings of the IEEE conference on computer vision and pattern recognition}, pages 2921--2929, 2016.

\end{thebibliography}
}

\appendix

\clearpage
\onecolumn
\begin{center}
    \large \bfseries Supplementary Material for From Visual Explanations to Counterfactual Explanations
with Latent Diffusion
\end{center}
\vspace{1em}

\setcounter{figure}{0}
\setcounter{table}{0}
\setcounter{equation}{0}

\section{Proof}
 
In this section, we provide the proofs for the propositions. Our objective is to assess the degree of deviation in optimization when replacing the gradient $\boldsymbol{u}^*$, which is the fastest direction, with the pruned gradient $\boldsymbol{v}^*$. Additionally, we establish the relationship between convergence and the thresholds $\xi_d$. We restate the terminology previously mentioned in the main paper: Suppose $\xi^*$ is the optimal threshold to split foreground $\mathcal{F}^*$ and background $\mathcal{B}^*$. $\mathcal{A} = \mathcal{F}\cup\mathcal{B}$ represents the set of all pixel coordinates in the original image $\mathbf{\mathcal{I}}^\textrm{F}$. First, $c: \mathbb{R}^{H_2\times W_2\times 1} \rightarrow \mathbb{R}^{H_2\times W_2\times C}$ expands latent mask by concatenating this binary mask itself $C$ times along the channel dimension. Second, let $g: \mathbb{R}^{H_2\times W_2 \times C} \rightarrow \mathbb{R}^{H_2W_2C}$ map the space $H_2\times W_2 \times C$ to a vector of dimension $H_2W_2C$, where $H_2, W_2, C$ is the height, width, channel of latent space. $\mathbf{\mathcal{M}}'$ represents the latent mask corresponding to the optimal threshold $\xi^*$.

\setcounter{prop}{0}
\begin{prop}
    \label{prop:proposition_1}
    Let $\boldsymbol{u}^* = g\left(\nabla_{\boldsymbol{z}^\textrm{init}}\mathcal{L}_{CE}\left(f_{cl}\left(\mathbf{\mathcal{I}}^\textrm{(k)}\right), \mathit{y}^\textrm{CF}\right)\right)$ is the optimal vector, and $\boldsymbol{v}^* = \boldsymbol{m}'\odot\boldsymbol{u}^*$ represents the pruned adversarial gradient vector with $\boldsymbol{m}'=g(c(\mathbf{\mathcal{M}}'))$. Then
    \begin{align}
        0^\circ\leq\ \angle (\boldsymbol{u}^*,\boldsymbol{v}^*) <90^\circ,
    \end{align}
    where $\boldsymbol{u}^*= (u_{1,1,1},\dots, u_{H_2,W_2,1},\dots, u_{H_2,W_2,4})$, $u_{i,j,s}$ represents the element corresponding to $\boldsymbol{\gamma}^\textrm{(k)}_{i,j,s}$. 
\end{prop}
\begin{proof}
    \begin{align}
        \cos(\angle\left(\boldsymbol{u}^*,\boldsymbol{v}^*\right)) & = \frac{\boldsymbol{u}^* \cdot \boldsymbol{v}^*}{\|\boldsymbol{u}^*\| \|\boldsymbol{v}^*\|} \nonumber\\
        & = \dfrac{\displaystyle\sum\limits_{s=1}^{4} \sum\limits_{(i,j) \in \mathcal{F}^*} u^2_{i,j,s}}{\underbrace{\sqrt{\sum\limits_{s=1}^{4} \sum\limits_{(i,j) \in \mathcal{A}} u_{i,j,s}^2}}_\text{constant}{\sqrt{\displaystyle\sum\limits_{s=1}^{4} \sum\limits_{(i,j) \in \mathcal{F}^*} u_{i,j,s}^2}}}\nonumber\\
        & = \gamma\sqrt{\sum\limits_{s=1}^{4} \sum\limits_{(i,j) \in \mathcal{F}^*} u_{i,j,s}^2} > 0 \quad (\gamma > 0). \nonumber
    \end{align}
    Obviously:
    \vspace*{-1.85pt}
    \begin{align}
        0^\circ\leq\angle\left(\boldsymbol{u}^*,\boldsymbol{v}^*\right)<90^\circ.\label{eq:9}
    \end{align}
\end{proof}

\setcounter{prop}{1}
\begin{prop}
    \label{prop:proposition_2}
    Suppose that $\{\xi_d\}^\infty_{d=1}$ is a sequence of thresholds satisfying:
    \begin{align}
        0\leq\xi_{d}<\xi_{d+1}< 1. \label{eq:3}
    \end{align}
    Then there exists a sequence of vectors $\left\{\boldsymbol{v}^{(d)}\right\}^\infty_{d=1}$ such that: 
    \begin{align}
        0^\circ\leq\angle \left(\boldsymbol{u}^*,\boldsymbol{v}^{(1)}\right)\leq\dots\leq\angle \left(\boldsymbol{u}^*,\boldsymbol{v}^{(\infty)}\right)<90^\circ. 
        \vspace{-1mm}
    \end{align}
\end{prop}
\begin{proof}
    For Equation~\ref{eq:3}, we have the foreground and background sets corresponding to each threshold $\xi$ that satisfy the following conditions: $\mathcal{F}_d \supseteq \mathcal{F}_{d+1}$, $\mathcal{B}_d \subseteq \mathcal{B}_{d+1}$, $\mathcal{A}=\mathcal{F}_d \cup \mathcal{B}_d$, where $d\in\mathbb{N}^*$. Thanks to Proposition~\ref{prop:proposition_1}, we continue to analyze:
    \vspace*{-1.85pt}
    \begin{align}
        \cos\left(\angle\left(\boldsymbol{u}^*, \boldsymbol{v}^{(d)}\right)\right) & = \gamma\sqrt{\sum\limits_{s=1}^{4} \sum\limits_{(i,j) \in \mathcal{F}_d} u_{i,j,s}^2}\\
        & \geq \gamma\sqrt{\sum\limits_{s=1}^{4} \sum\limits_{(i,j) \in \mathcal{F}_{d+1}} u_{i,j,s}^2}\\
        & =\cos\left(\angle\left(\boldsymbol{u}^*, \boldsymbol{v}^{(d+1)}\right)\right)
    \end{align}
    Therefore:
    \vspace{-4pt}
    \begin{align}
        0^\circ\leq\angle\left(\boldsymbol{u}^*, \boldsymbol{v}^{(d)}\right) \leq \angle\left(\boldsymbol{u}^*, \boldsymbol{v}^{(d+1)}\right)\leq 90^\circ, \forall d\in\mathbb{N}^*.
    \end{align}
    This proof is complete.
\end{proof}

\section{Overview of ECED}
In this section, we provide the following information and material.
\begin{itemize}
    \item Algorithms of our ECED.
    \item The implementation details and hyperparameter configurations of Latent Diffusion.
    \item Fine-tuning strategy for Stable Diffusion on the CelebA-HQ dataset.
    \item The ability to leverage the context of Latent Diffusion.
    \item The efficiency of blending latents.
    \item The preservation strategy of our method.
    \item More qualitative results.
\end{itemize}
\subsection{Algorithms}
\begin{algorithm}
\caption{Identifying key region}\label{algo:step1}
    \begin{algorithmic}[1]
        \Require Initial image $\mathbf{\mathcal{I}}^\textrm{F}$, original label $\mathit{y}^\textrm{F}$, pretrained classifier $f_{cl}$, classifier's specified layer $l$, threshold $\xi$ spliting two parts of the image, visual explanation algorithm $ScoreCAM$
        \Function{IDENTIFY-FG}{$\mathbf{\mathcal{I}}^\textrm{F}$, $\mathit{y}^\textrm{F}$, $l$, $\xi$}
        \State $\mathbf{U} = ScoreCAM(\mathbf{\mathcal{I}}^\textrm{F}, \mathit{y}^\textrm{F}, f_{cl}, l)$
        \Comment{Extract the attention map}
        \State $\mathbf{\mathcal{M}} = \mathbf{U}[u_{i,j}>\xi]$ 
        \Comment{Get the binary mask}
        \State $\mathbf{\mathcal{M}}' = downsample(\mathbf{\mathcal{M}})$
        \Comment{Get the latent mask}
        \State \Return $\mathbf{\mathcal{M}}'$
        \EndFunction
    \end{algorithmic}
\end{algorithm}
\begin{algorithm}
\caption{Preserving background during image synthesis}\label{algo:step2}
    \begin{algorithmic}[1]
        \Require Initial image $\mathbf{\mathcal{I}}^\textrm{F}$, latent mask $\mathbf{\mathcal{M}}'$, $VAE = (\mathcal{E}(\cdot),\mathcal{D}(\cdot))$, noise coefficient $\alpha_0$, distance loss $\mathcal{L}_{dis}$, number of update iterations $N$, optimization algorithm $Adam$
        \Function{PRESERVE-BG}{$\mathbf{\mathcal{I}}^\textrm{F}$, $\mathbf{\mathcal{M}}'$}
        \State $\boldsymbol{z}^\textrm{F}=\boldsymbol{z}^\textrm{bg}=\mathcal{E}(\mathbf{\mathcal{I}}^\textrm{F})$
        \Comment{Init latents}
        \For{$i = 1,\dots,N$}
            \State $\boldsymbol{\epsilon} \sim \mathcal{N}(\mathbf{0},\mathbf{I})$
            \State $\boldsymbol{z}^\textrm{init} \leftarrow \boldsymbol{z}^\textrm{F}\odot\mathbf{\mathcal{M}}'+\boldsymbol{z}^\textrm{bg}\odot(\mathbf{1}-\mathbf{\mathcal{M}}')$
            \Comment{Blend latents}
            \State $\boldsymbol{z}_0 \leftarrow \sqrt{\alpha_0}\boldsymbol{z}^\textrm{init}+\sqrt{1-\alpha_0}\boldsymbol{\epsilon}$
            \Comment{Add noise}
            \State $\mathbf{\mathcal{I}}' \leftarrow \mathcal{D}(\boldsymbol{z}_0)$
            \State $grad_1 \leftarrow \nabla_{\boldsymbol{z}^\textrm{bg}}\mathcal{L}_{dis}(\mathbf{\mathcal{I}}',\mathbf{\mathcal{I}}^\textrm{F}, \mathbf{\mathcal{M}}')$
            \State $grad_2 \leftarrow \nabla_{\theta_\mathcal{D}}\mathcal{L}_{dis}(\mathbf{\mathcal{I}}',\mathbf{\mathcal{I}}^\textrm{F}, \mathbf{\mathcal{M}}')$
            \State $\boldsymbol{z}^\textrm{bg} \leftarrow Adam(\boldsymbol{z}^\textrm{bg}, grad_1)$
            \Comment{Update}
            \State $\theta_\mathcal{D} \leftarrow Adam(\theta_\mathcal{D}, grad_2)$
            \Comment{Update}
        \EndFor
        \State \Return $\boldsymbol{z}^\textrm{bg*},\theta_\mathcal{D}^*$
        \EndFunction
    \end{algorithmic}
\end{algorithm}
\begin{algorithm}
\caption{Generating counterfactual explanation}\label{algo:step3}
    \begin{algorithmic}[1]
        \Require Initial image $\mathbf{\mathcal{I}}^\textrm{F}$, target label $\mathit{y}^\textrm{CF}$, latent mask $\mathbf{\mathcal{M}}'$, original latent $\boldsymbol{z}^\textrm{F}$, optimal background latent $\boldsymbol{z}^\textrm{bg*}$, Latent Diffusion model $SD=\{(\mathcal{E}(\cdot),\mathcal{D}^*(\cdot)), DiffusionModel =(noise(\boldsymbol{z},t), denoise(\boldsymbol{z},\mathbf{C},t))\}$, text encoder CLIP, sequence of noise coefficients $\{\alpha_t\}^{T_1}_{t=0}$, diffusion steps $\tau$, classifier $f_{cl}$, Cross-Entropy loss $\mathcal{L}_{CE}$, number of update iterations $T_2$, optimization algorithm $Adam$
        \Function{GENERATE-CE}{$\mathbf{\mathcal{I}}^\textrm{F}$, $\mathit{y}^\textrm{CF}$, $\mathbf{\mathcal{M}}'$, $\boldsymbol{z}^\textrm{bg*}$}
            \State $\mathbf{C}^\textrm{CF} = CLIP(\mathit{y}^\textrm{CF})$
            \State $\boldsymbol{z}^\textrm{init} \leftarrow \boldsymbol{z}^\textrm{F}\odot\mathbf{\mathcal{M}}'+\boldsymbol{z}^\textrm{bg*}\odot(\mathbf{1}-\mathbf{\mathcal{M}}')$
            \LineComment{Attack iteration steps}
            \For{$t_2=0,\dots, T_2$}
                \LineComment{Blended latent diffusion algorithm}
                \State $\boldsymbol{z}_\tau \sim noise(\boldsymbol{z}^\textrm{init},\tau)$
                \For{$t=\tau,\dots,0$}
                    \State $\boldsymbol{z}^\textrm{fg} \sim denoise(\boldsymbol{z}_t, \mathbf{C}^\textrm{CF}, t)$
                    \State $\boldsymbol{z}^\textrm{bg*}_t \sim noise(\boldsymbol{z}^\textrm{bg*},t)$
                    \State $\boldsymbol{z}_t \leftarrow \boldsymbol{z}^\textrm{fg}\odot\mathbf{\mathcal{M}}'+\boldsymbol{z}^\textrm{bg*}_t\odot(\mathbf{1}-\mathbf{\mathcal{M}}')$
                \EndFor
                \State $\mathbf{\mathcal{I}}^{(t_2)}\leftarrow\mathcal{D}^*(\boldsymbol{z}_0)$
                \State $grad \leftarrow \nabla_{\boldsymbol{z}^\textrm{init}}\mathcal{L}_{CE}\left(f_{cl}\left(\mathbf{\mathcal{I}}^{(t_2)}\right), \mathit{y}^\textrm{CF}\right)$
                \State $\boldsymbol{z}^\textrm{init}\leftarrow Adam(\boldsymbol{z}^\textrm{init}, grad\odot\mathbf{\mathcal{M}}')$ 
                \Comment{Update with pruning-based attack}
            \EndFor
            \State $\mathbf{\mathcal{I}}^\textrm{CF} = \mathbf{\mathcal{I}}^{(T_2)}$
            \Comment{Counterfactual explanation}
        \State \Return $\mathbf{\mathcal{I}}^\textrm{CF}$
        \EndFunction
    \end{algorithmic}
\end{algorithm}

\clearpage
\subsection{Implementation Details}
In this work, we implement the blended latent diffusion algorithm proposed in \cite{avrahami2023blended}. To reiterate, this algorithm blends latent representations at each timestep $t$, defined as follows:
\begin{align}
    \boldsymbol{z}^\textrm{bg}_t &= \sqrt{\bar\alpha_t}\boldsymbol{z}^\textrm{F} + \sqrt{1 - \bar\alpha_t}\boldsymbol{\epsilon}_t, \label{eq:1}\\
    \boldsymbol{z}^\textrm{fg}_{t} & \approx \sqrt{\alpha_{t-1}}\hat{\boldsymbol{z}}_0 + \beta_{t-1}\boldsymbol{\epsilon}_\theta(\boldsymbol{z}_t, \mathbf{C}^\textrm{CF}) + \sigma_t\boldsymbol{\epsilon}_t.\label{eq:2}
\end{align}
According to the hyperparameter settings and configuration of Stable Diffusion, $\{ \beta_d \}^{T_1}_{t=0}$ defines a linear noise scheduling with $\beta_0 = 0.00085$ and $\beta_{T_1} = 0.012$ ($\alpha_t = 1$ if $t<0$), and $\bar\alpha_t = \prod_{s=0}^t \alpha_s$. The representations and coefficients in Equation~\ref{eq:1} are: $\boldsymbol{\epsilon}_t \sim\mathcal{N}(0,\textbf{I})$, $\hat{\boldsymbol{z}_0} = \frac{\boldsymbol{z}_t - \sqrt{1-\bar\alpha_t}\boldsymbol{\epsilon}_\theta(\boldsymbol{z}_t, \mathbf{C}^\textrm{CF})}{\sqrt{\bar\alpha_t}}$, $\beta_{t} = \sqrt{1 - \alpha_{t-1} - \sigma_t^2}$, $\sigma_t = \sqrt{\frac{1 - \alpha_{t-1}}{1 - \alpha_t}} \sqrt{1 - \frac{\alpha_t}{\alpha_{t-1}}}$.

To represent the target classes, these conditions are mapped to CLIP-style text prompts \cite{radford2021learning}, as follows:
\begin{itemize}
    \item ImageNet: \texttt{A photo of a/an \{category\}.} (\texttt{attribute} $\in$ \{cougar, cheetah, sorrel, zebra, Persian cat, Egyptian cat\}).
    \item CelebA-HQ: \texttt{A photo of a/an \{attribute\} face.} (\texttt{attribute} $\in$ \{smiling, non smiling, young, old\}).
\end{itemize}
For CelebA-HQ dataset, we fine-tuned the Stable Diffusion model to align the generated images with the desired conditions. Additionally, the purpose of this optimization is to generate a set of images that closely resemble the data distribution, thereby improving the FID score. Specifically, we optimized the weights of the UNet \cite{ronneberger2015u} to minimize the following loss:
\begin{align}
    \mathcal{L}_{LDM} = \mathbb{E}_{\mathcal{E}(\mathbf{x}), \boldsymbol{\epsilon}\sim\mathcal{N}(0,\mathbf{I}),t}\left[\lVert\boldsymbol{\epsilon}-\boldsymbol{\epsilon}_\theta(\boldsymbol{z}_t, \mathbf{C}, t)\rVert^2_2\right].
\end{align}
We utilized the AdamW optimizer \cite{loshchilov2017decoupled} with $betas=(0.9, 0.999)$ and a learning rate of $10^{-4}$.
\subsection{Blending Approach in Latent Diffusion}
\begin{figure}
    \centering
    \includegraphics[width=1.0\linewidth]{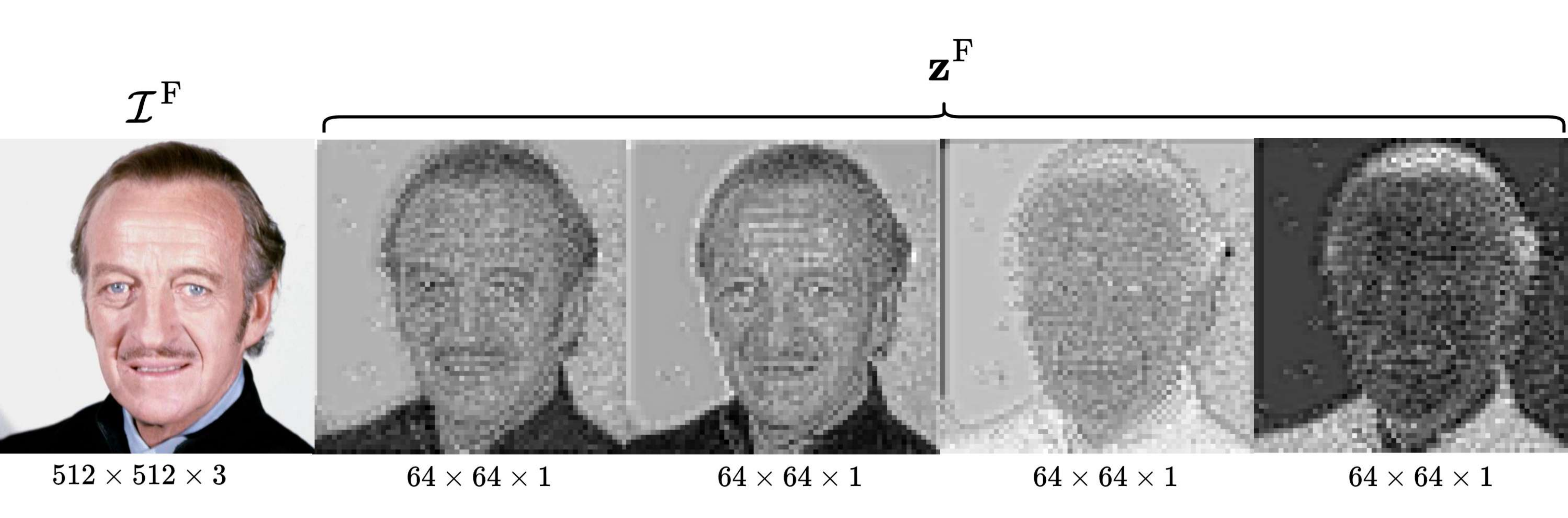}
    \caption{The visualization of the latent space.}
    \label{fig:latent}
\end{figure}

We provide examples in Figure~\ref{fig:latent}. The channels of the latent variable capture high-level features of the original image. Therefore, Avrahami~\textit{et al.}'s approach~\cite{avrahami2023blended} effectively separated the latent subspaces of background and foreground using a latent mask rescaled from the original binary mask. 
Combined with the optimal latent $\boldsymbol{z}^\textrm{bg*}$, the details in the counterfactual image are likely to tightly align the context. This is primarily because Stable Diffusion integrated two attention mechanisms into the UNet model during the denoising process. Specifically, self-attention highlights the importance of latent features relative to the query feature, while cross-attention indicates the significance of positions on the generated image with respect to the content of the text prompt. These issues have been discussed in related works \cite{liu2024towards,hertz2022prompt}.
\subsection{Preservation Strategy}
Based on the blending strategy mentioned in Algorithm~\ref{algo:step3}, we observed that the latent variable $\boldsymbol{z}_t$ retains a small amount of noise in the subspace related to the background at timestep $t=0$, corresponding to the noise coefficient $\beta_0 = 0.00085$. This is why we simulate the noise addition process before decoding the latent $\boldsymbol{z}_0$ back to the pixel space in the second phase. In the experimental section, we verified the effectiveness of this approach by calculating the pixel-wise difference in the background region between the reconstructed image and the original image. Formally, we calculate the loss as follows:
\begin{align}
    \mathcal{L}_{bg} = \frac{1}{|\mathcal{B}|}\sum_{(i,j)\in \mathcal{B}}MSE(\mathcal{D}(\boldsymbol{z}_0)_{i,j}, \mathcal{I}^\textrm{F}_{i,j}),
\end{align}
where MSE denotes the Mean Squared Error, and $\mathcal{B}$ represents the locations of the pixels in the background. We conducted experiments and computed the average difference over 100 random samples, and then obtained the confidence interval based on the normal distribution.

\subsection{More qualitative results}
\begin{figure*}
    \centering
    \includegraphics[width=1.0\linewidth]{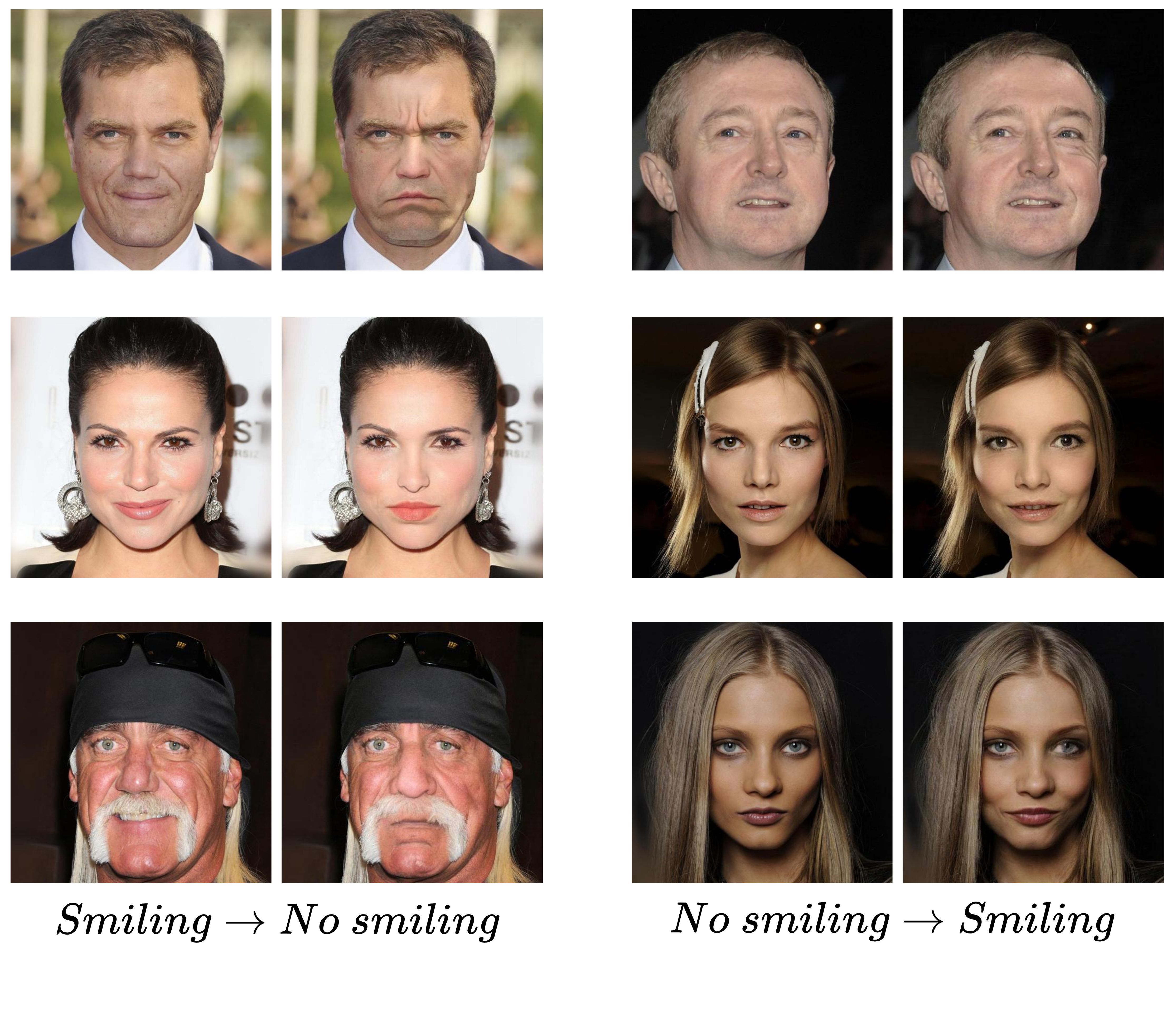}
    \caption{\textbf{Qualitative results for `Smile' attribute with VGG-16.} Left to right: original image, counterfactual image generated by ECED.}
    \label{fig:supp_smile}
\end{figure*}

\begin{figure*}
    \centering
    \includegraphics[width=1.0\linewidth]{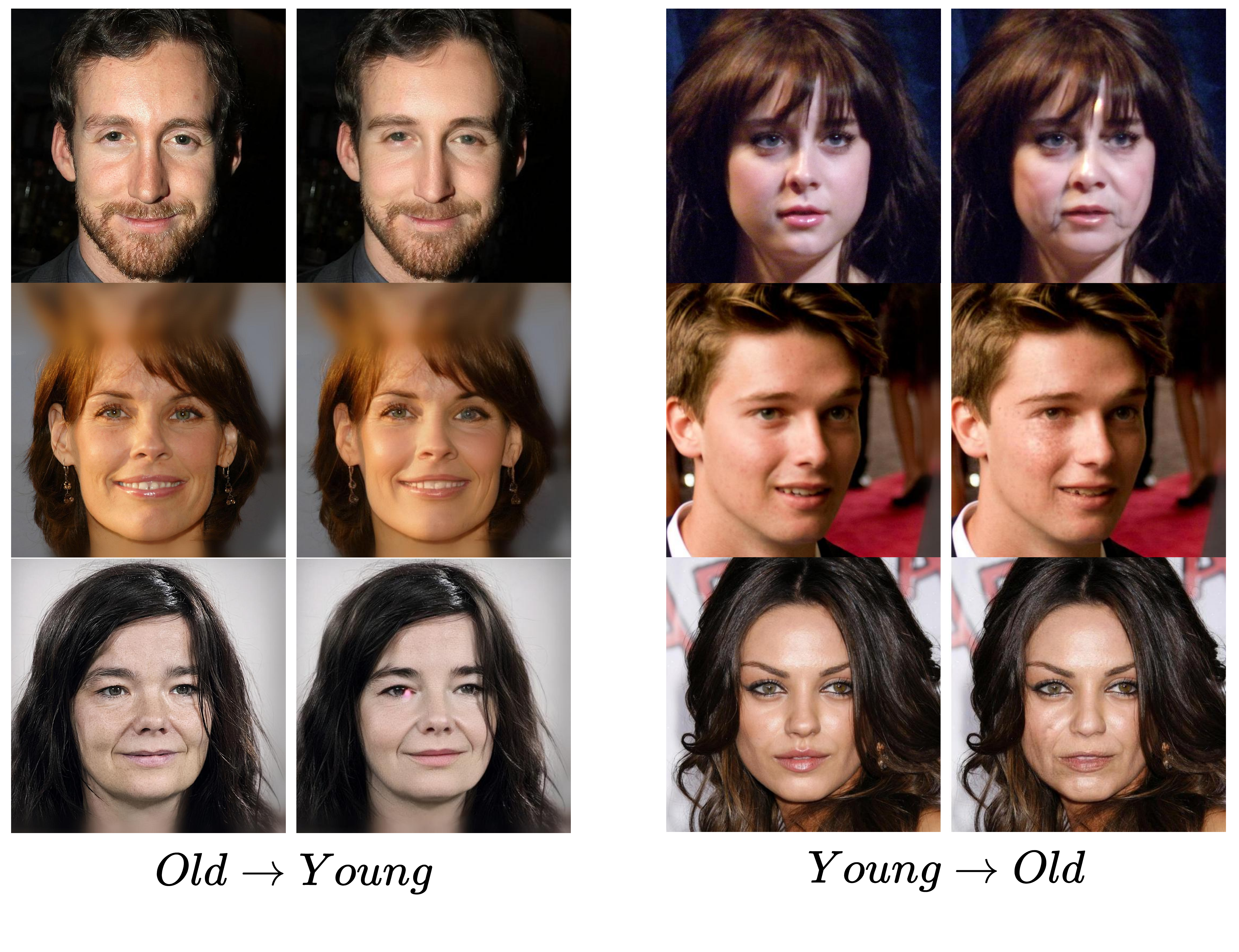}
    \caption{\textbf{Qualitative results for `Age' attribute with VGG-16.} Left to right: original image, counterfactual image generated by ECED.}
    \label{fig:supp_age}
\end{figure*}

\begin{figure}
    \centering
    \includegraphics[width=1.0\linewidth]{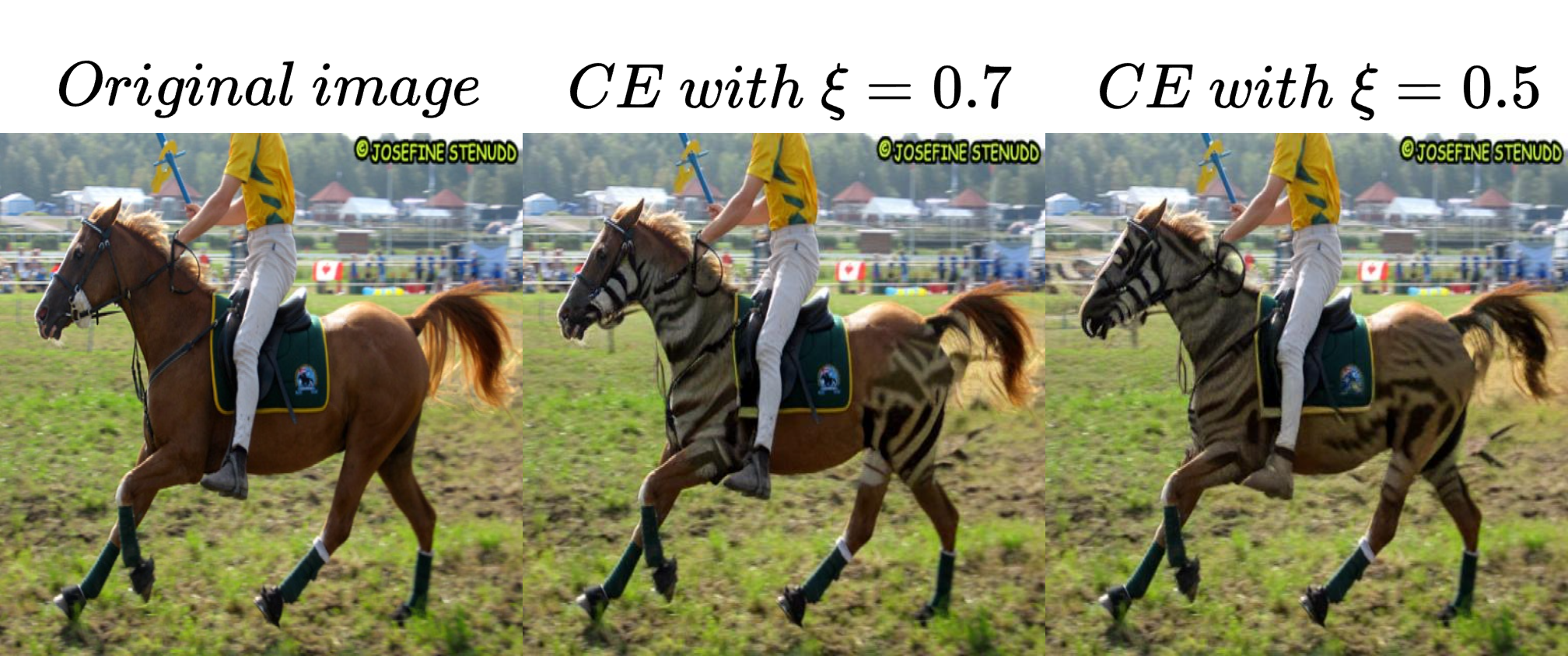}
    \caption{The example of diversity in generating counterfactual explanations by ECED.}
    \label{fig:diversity}
\end{figure}

 We provide counterfactual explanations, presented in Figure~\ref{fig:supp_smile} and ~\ref{fig:supp_age}. Additionally, we examine the diversity in generating CEs by ECED by setting different thresholds $\xi$, as shown in Figure~\ref{fig:diversity}.
\section{Evaluation Protocols for Counterfactual Explanations}
Visual counterfactual explanations are evaluated based on three key criteria: Closeness/Sparsity, Validity, and Realism.
\subsection{Sparsity}
\textit{Euclidean distance} matches the $p$-th order discrepancy of pixel values at each corresponding position between the original image $\mathcal{I}^\textrm{F}$ and the counterfactual image $\mathcal{I}^\textrm{CF}$:
\begin{align}
    \mathcal{L}_{p} & = \frac{1}{N}\sum_{i=1}^{N} \| d_{i} \|_{p} \\
                    & = \frac{1}{N}\sum_{i=1}^{N}(\sum_{c=1}^{C} \sum_{h=1}^{H_1} \sum_{w_=1}^{W_1} |\mathbf{\mathcal{I}}_{i,c,h,w}^\textrm{F} - \mathbf{\mathcal{I}}_{i,c,h,w}^\textrm{CF}|^{p})^{\frac{1}{p}},
\end{align}
where $N$ is the number of images, and $C, H_1, W_1$ are the number of channels, height, and width of the images, respectively, with $p > 0$.

\textit{SimSiam Similarity} measures the cosine similarity between the counterfactual image $\mathcal{I}^\textrm{CF}$ and the corresponding original image $\mathcal{I}^\textrm{F}$ in the feature space extracted by the self-supervised $SimSiam$ model \cite{chen2021exploring}.
\begin{equation}
S^3(\mathcal{I}^\textrm{CF}, \mathcal{I}^\textrm{F}) = \frac{{ \mathcal{S}(\mathcal{I}^\textrm{CF}) \cdot  \mathcal{S}(\mathcal{I}^\textrm{F})}}{{\lVert  \mathcal{S}(\mathcal{I}^\textrm{CF}) \rVert \lVert \mathcal{S}(\mathcal{I}^\textrm{F}) \rVert}}. 
\end{equation}

\textit{Correlation Difference} (CD) and \textit{Mean Number of Attribute Changes} (MNAC) measure the average number of attributes modified in the counterfactual explanation, where MNAC addresses the limitations of CD.
\begin{align}
\text{MNAC}
=\frac{1}{N} \sum_{i=1}^{N} \sum_{a \in \mathcal{A} }^{} \left[ \mathbb{I}\left( \mathbb{I}\left( O_a(\mathcal{I}^{CF}_{i}) > \beta \right) \neq \mathbb{I}\left( O_a(\mathcal{I}^{F}_i) > \beta \right) \right) \right],
\end{align}

\begin{equation}
\textrm{CD}_q = \frac{1}{N} \sum_{i=1}^{N} \sum_{a \in \mathcal{A} }^{}  |c^{q,a}(\mathcal{I}^{CF}_i)-c^{q,a}(\mathcal{I}^{F}_i)|.
\end{equation}

\textit{Counterfactual Transition} (COUT) \cite{khorram2022cycle} measures the sparsity of changes in counterfactual explanations. It quantifies the impact of perturbations applied to the factual image $\mathcal{I}^\textrm{F}$ by using a normalized mask $m = \delta(||x^{F}-x^{CF}||1)$ that represents the relative change compared to the counterfactual image, where $\delta$ normalizes the absolute values to the range $[0,1]$. The computation of COUT is performed incrementally by gradually inserting the highest-ranked pixel groups from $\mathcal{I}^\textrm{CF}$ based on these sorted mask values.

At each step of adding pixel groups \begin{math}t \in \{0, \ldots, T\}\end{math}, the measure calculates the probability $\mathit{f}_{cl}(\cdot)$ for the original label and the desired label, $\mathit{y}^\textrm{F}$ and $\mathit{y}^\textrm{CF}$, through the transition from $x_0 = \mathcal{I}^\textrm{F}$ to $x_T = \mathcal{I}^\textrm{CF}$. From this, the COUT score is defined as:
\begin{equation}
\textrm{COUT} = \textrm{AUPC}(\mathit{y}^\textrm{CF}) - \textrm{AUPC}(\mathit{y}^\textrm{F})\in[-1,1].
\end{equation}
The perturbation area under the curve for each label \begin{math}y \in {\mathit{y}F, \mathit{y}{CF}}\end{math} is calculated as follows:
\begin{align}
\textrm{AUPC}(y) &= \frac{1}{T} \sum_{t=0}^{T-1} \frac{1}{2} \left(f_{cl}\left(x_t, \mathit{y}\right) + f_{cl}\left(x_{t+1}, \mathit{y}\right)\right) \\
\textrm{AUPC}(y)&\in [0,1].
\end{align}
\subsection{Validity}
\textit{Flip Ratio} (FR) measure is commonly used to assess the authenticity of counterfactual outcomes for the desired label. This criterion focuses on evaluating the validity of $N$ counterfactual explanations by measuring the extent to which the original label $\mathit{y}^\textrm{F}_i$ of the $i$-th original image $\mathcal{I}^{\textrm{F},i}$ shifts the classification model’s prediction to the counterfactual target class $\mathit{y}^\textrm{CF}_i$ for the counterfactual image $\mathcal{I}^{\textrm{CF},i}$.
\begin{equation}
FR = \frac{\sum\limits_{i=1}^N \mathbb{I}\left(\mathit{f}_{cl}(\mathcal{I}^{\textrm{CF},i}) = \mathit{y}^\textrm{CF}_i\right)}{N},
\end{equation}
where $\mathbb{I}$ is the indicator function.
\subsection{Realism}
\textit{Fréchet Inception Distance} (FID) assesses the realism of generated images by measuring the FID distance between the distributions of features (extracted via the {InceptionV3} network \cite{szegedy2016rethinking}) in the original dataset and the counterfactual image set:
\begin{equation}
\begin{aligned}
\text{FID} &= \|\mu_{F} - \mu_{CF}\|_2^2 + \text{Tr}(\Sigma_{F} + \Sigma_{CF} - 2\sqrt{\Sigma_{F}\Sigma_{CF}}),
\end{aligned}
\end{equation}
where $\mu_{F}, \mu_{CF}$ represent the mean vectors, and $\Sigma_{F}, \Sigma_{CF}$ represent the covariance matrices derived from the feature distributions of the {InceptionV3} model for the real and counterfactual image sets, respectively. However, due to the nature of this type of explanation, which creates very subtle changes (differing only at a few pixels), this introduces significant bias. To address this issue, \cite{jeanneret2023adversarial} proposed a solution by splitting the dataset to compute {FID}, termed {sFID}. The basic idea is to split both the real and counterfactual image sets into two subsets, then compute the {FID} value across the cross subsets (i.e., real image subsets not paired with their respective real counterparts). Finally, the average of the two {FID} values is taken to obtain the {sFID} value.

\end{document}